\documentclass[10pt,twocolumn,letterpaper]{article}

\usepackage{wacv}
\usepackage{times}
\usepackage{epsfig}
\usepackage{graphicx}
\usepackage{amsmath}
\usepackage{amssymb}
\usepackage{bm}

\wacvfinalcopy 

{\def\wacvPaperID{95} 
{}

{\ifwacvfinal\pagestyle{empty}\fi}
{\setcounter{page}{1}}
\begin{document}

{\title{NeurReg: Neural Registration and Its Application to Image Segmentation}}


{\author{Wentao Zhu \hspace{0.5cm} Andriy Myronenko \hspace{0.5cm} Ziyue Xu \hspace{0.5cm} Wenqi Li \\ Holger Roth \hspace{0.5cm} Yufang Huang{$^{2}$} \hspace{0.5cm} Fausto Milletari \hspace{0.5cm} Daguang Xu \\
{$^1$}NVIDIA \hspace{0.5cm} {$^2$}Cornell University\\
{{\tt\small {{wentaoz,amyronenko,ziyuex,wenqil,hroth,fmilletari,daguangx}@nvidia.com}}} \\ {{\tt\small {yfhuang1992new@gmail.com}} }
}}

{\maketitle}
{\ifwacvfinal\thispagestyle{empty}\fi}
\begin{abstract}
   Registration is a fundamental task in medical image analysis which can be applied to several tasks including image segmentation, intra-operative tracking, multi-modal image alignment, and motion analysis. Popular registration tools such as ANTs and NiftyReg optimize an objective function for each pair of images from scratch which is time-consuming for large images with complicated deformation. Facilitated by the rapid progress of deep learning, learning-based approaches such as VoxelMorph have been emerging for image registration. These approaches can achieve competitive performance in a fraction of a second on advanced GPUs. In this work, we construct a neural registration framework, called NeurReg, with a hybrid loss of displacement fields and data similarity, which substantially improves the current state-of-the-art of registrations. Within the framework, we simulate various transformations by a registration simulator which generates fixed image and displacement field ground truth for training. Furthermore, we design three segmentation frameworks based on the proposed registration framework: 1) atlas-based segmentation, 2) joint learning of both segmentation and registration tasks, and 3) multi-task learning with atlas-based segmentation as an intermediate feature. Extensive experimental results validate the effectiveness of the proposed NeurReg framework based on various metrics: the endpoint error (EPE) of the predicted displacement field, mean square error (MSE), normalized local cross-correlation (NLCC), mutual information (MI), Dice coefficient, uncertainty estimation, and the interpretability of the segmentation. The proposed NeurReg improves registration accuracy with fast inference speed, which can greatly accelerate related medical image analysis tasks. 
\end{abstract}

\section{Introduction}
Image registration tries to establish the correspondence between objects, edges, surfaces or landmarks in different images and it is critical to many clinical tasks such as image fusion, organ atlas creation, and tumor growth monitoring~\cite{haskins2019deep}. Manual image registration is laborious and lacks reproducibility which causes potentially clinical disadvantage. Therefore, automated registration is desired in many clinical settings. Generally, registration can be necessary to analysis sequential data~\cite{zhu2016co} or a pair of images from different modalities, acquired at different times, from different viewpoints or even from different patients. Thus designing a robust image registration can be challenging due to the high variability.

Traditional registration methods are based on estimation of the displacement field by optimizing certain objective functions. Such displacement field can be modeled in several ways, e.g. elastic-type models~\cite{bajcsy1989multiresolution,shen2002hammer}, free-form deformation (FFD)~\cite{rueckert1999nonrigid}, Demons~\cite{thirion1998image}, and statistical parametric mapping~\cite{ashburner2000voxel}. Beyond the deformation model, diffeomorphic transformations~\cite{krebs2019learning} preserve topology with exact inverse transforms and many methods adopt them such as LDDMM~\cite{beg2005computing}, SyN~\cite{avants2008symmetric} and DARTEL~\cite{ashburner2007fast}. One limitation of these methods is that the optimization can be computationally expensive.

Deep learning-based registration methods have recently been emerging as a viable alternative to the above conventional methods~\cite{rohe2017svf,sokooti2017nonrigid,yang2017quicksilver}. These methods employ sparse/weak label of registration field, or conduct supervised learning purely based on registration field, inducing high sensitivity on registration field during training. Recent unsupervised deep learning-based registrations, such as VoxelMorph~\cite{balakrishnan2018unsupervised}, are facilitated by a spatial transformer network~\cite{jaderberg2015spatial,de2017end,dalca2018unsupervised}. VoxelMorph is also further extended to diffeomorphic transformation and Bayesian framework~\cite{dalca2018unsupervised}. NMSR employs self-supervised optimization and multi-scale registration to handle domain shift and large deformation~\cite{zhu2019neural}. However from the performance perspective, most of these registration methods have comparable accuracy as traditional iterative optimization methods, although with potential speed advantages. 

In this work, we design a deep learning-based registration framework with a hybrid loss based on data similarity and registration field supervision. The framework, as illustrated in Fig.~\ref{fig:framework}, is motivated by the fact that supervised learning can predict accurate displacement field, while unsupervised learning can extract visual representations that generalize well to unseen images. More specifically, we build a registration simulator which models random translation, rotation, scale, and elastic deformation. For training, we employ the registration simulator to generate a fixed image with its corresponding displacement field ground truth from a given moving image. Then, we use a U-Net to parameterize the displacement field and a spatial transform network to warp the moving image towards the generated fixed image~\cite{ronneberger2015u,jaderberg2015spatial}. In addition to the hybrid loss and registration simulator for registration itself, we further investigated its potential in segmentation, a common application of deformable registration. We design two different multi-task learning networks between segmentation and registration. Also, to fully exploit the capacity with one moving/atlas image, we further design a dual registration to enforce registration loss and segmentation loss from both a random training image and a random image from registration simulator. 

{\small \begin{figure*}[t]
\begin{center}
   \includegraphics[width=0.9\linewidth]{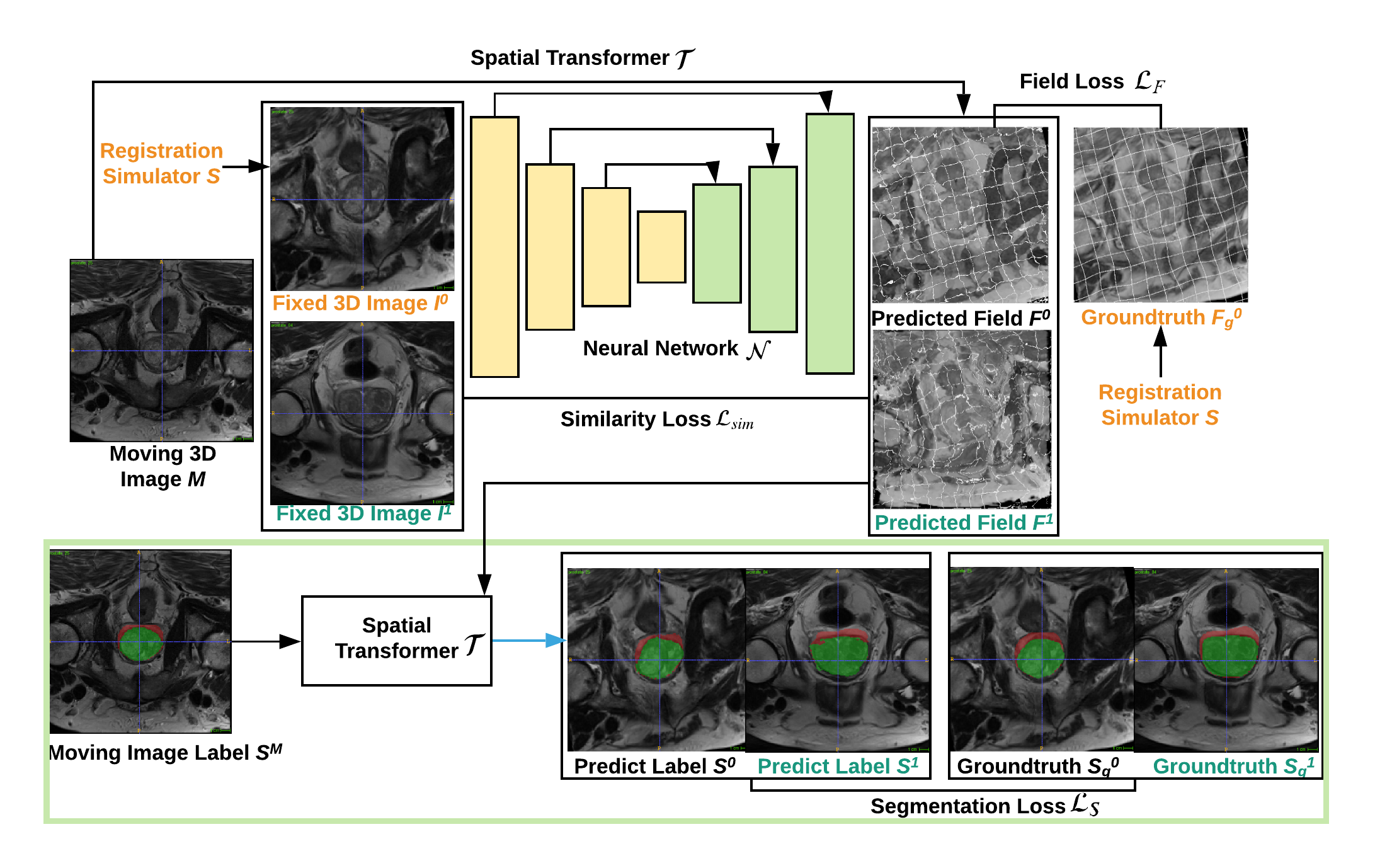}
\end{center}
   \caption{Framework of the proposed NeurReg. In the registration module, we generate a random simulated displacement field and fixed image given a moving image. During training, a hybrid loss consisting of displacement field loss and data similarity loss is employed to take advantage of accurate field supervision and powerful generalization of appearance similarity. For the registration-based segmentation, a dual registration scheme is designed with multi-task loss for registration and segmentation. The dual registration enforces the moving/atlas image to be aligned with random images from both the registration simulator and the training set.}
\label{fig:framework}
\end{figure*}}

Our main contributions are as follows: 1) we design a registration simulator to model various transformations, and employ a hybrid loss with registration field supervision loss and data similarity loss to benefit from both accurate supervision and powerful generalization of deep appearance representation. 2) We design a dual registration scheme in the multi-task learning between registration and segmentation to fully exploit the capacity of one moving/atlas image. We expect the network to align the moving image with a pair of random images from the registration field and training set. A residual segmentation block is further developed to boost the performance as illustrated in Fig.~\ref{fig:boost_segm}. 3) We validate our framework on two widely used public datasets: the Hippocampus and Prostate datasets from the Medical Segmentation Decathlon~\cite{simpson2019large}. Our framework outperforms three popular public toolboxes, ANTs~\cite{avants2009advanced}, NiftyReg~\cite{modat2010fast} and VoxelMorph~\cite{balakrishnan2018unsupervised}, on both registration and segmentation tasks based on several evaluation metrics. 

\section{Related Work}
Deep learning-based medical image registration can be primarily categorized into three classes, deep iterative registration, supervised, and unsupervised transformation estimation~\cite{haskins2019deep}. Early deep learning-based registrations directly embed deep learning as a visual feature extractor into traditional iterative registration based on hand-crafted metrics such as sum of squared differences (SSD), cross-correlation (CC), mutual information (MI), normalized cross correlation (NCC) and normalized mutual information (NMI). Wu et al.~\cite{wu2013unsupervised} employed a stacked convolutional auto-encoder to extract features for mono-modal deformable registration based on NCC. Blendowski et al.~\cite{blendowski2019combining} combine CNN feature with MRF-based feature in mono-modal registration. Reinforcement learning is further used to mimic the iterative transformation estimation. Liao et al.~\cite{liao2017artificial} use reinforcement learning and a greedy supervision to conduct rigid registration. Kai et al.~\cite{ma2017multimodal} further use Q-learning and contextual feature to perform rigid registration. Miao et al.~\cite{miao2018dilated} then employ multi-agent-based reinforcement learning in the rigid registration. Krebs et al.~\cite{krebs2017robust} conduct deformable registration by reinforcement learning on low resolution deformation with fuzzy action control. The iterative approaches can be relatively slow compared with the direct transformation estimation.

Supervised transformation uses a neural network to estimate transformation parameters directly which can significantly speed up the registration. AIRNet uses a CNN to directly estimate rigid transformation~\cite{chee2018airnet}. Rothe et al.~\cite{rohe2017svf} use a U-Net to estimate the deformation field. Cao et al.~\cite{cao2017deformable} perform displacement field estimation based on image patches and an equalized activate-points guided sampling is proposed to facilitate the training. Sokooti et al.~\cite{sokooti2017nonrigid} employ random deformation field to augment the dataset and design a multi-scale CNN to predict a deformation field. Uzunova et al.~\cite{uzunova2017training} use a CNN to fit registration field from statistical appearance models. The supervised transformation estimation heavily relies on the quality/diversity of registration field ground truth generated synthetically or manually from expert. Unsupervised registration is desirable to learn representation from data to increase generalization.

Unsupervised transformation estimation mainly uses spatial transformer networks (STN) to warp moving image with estimated registration field, and training of the estimators relies on the design of data similarity function and smoothness of estimated registration field~\cite{jaderberg2015spatial}. Neylon et al.~\cite{neylon2017neural} model the relationship between similarity function and TRE. VoxelMorph is designed as a general method for unsupervised registration and further extended to variational inference for deformation field~\cite{dalca2018unsupervised,balakrishnan2018unsupervised,balakrishnan2019voxelmorph}. Adversarial similarity network further employs a discriminator to automatically learn the similarity function~\cite{fan2018adversarial}. Jiang et al.~\cite{jiang2018cnn} instead learn parameterization of multi-grid B-Spline by a CNN. NMSR employs self-supervised optimization and multi-scale registration to handle domain shift and large deformation~\cite{zhu2019neural}. These unsupervised transformation estimations produce comparable accuracy and are significantly faster than the traditional registration tools~\cite{balakrishnan2018unsupervised,dalca2018unsupervised}. 

Different from the aforementioned methods, we design a neural registration, NeurReg, by taking advantage of both accurate supervision on registration field and good generalization of unsupervised similarity objective function. For registration-based segmentation, a dual registration framework is proposed by enforcing the moving/atlas image to align with random images from both training set and registration simulator which is capable to model various transformations. The NeurReg obtains better performance than ANTs~\cite{avants2009advanced}, NiftyReg~\cite{modat2010fast} and VoxelMorph~\cite{balakrishnan2018unsupervised} and can be a new baseline for medical image registration.

\section{Neural Registration Framework}
In this section, we introduce three main components in the neural registration framework, registration simulator, neural registration and registration-based segmentation, as illustrated in Fig.~\ref{fig:framework}. 

\subsection{Registration Simulator}\label{sec:rs}
Given a moving/atlas image $\bm{M}$, we generate a random deformation $\bm{F}_g^0$ and a fixed image $\bm{I}^0$ from a registration simulator $\mathcal{S}$. The generated random deformation field $\bm{F}_g^0$ can be used as an accurate supervised loss $\mathcal{L}_{F}$ for predicted registration field $\bm{F}^0$ parameterized by a neural network $\mathcal{N}$. To fully learn representation from data, a similarity loss $\mathcal{L}_{sim}$ is employed to measure the discrepancy between the fixed image $\bm{I}_0$ and warped image $\mathcal{T}(\bm{M}, \bm{F}^0)$ through spatial transformer $\mathcal{T}$~\cite{jaderberg2015spatial}.

To fully model various transformations, we simulate random rotation, scale, translation and elastic deformation in the registration simulator $\mathcal{S}$~\cite{simard2003best}. We uniformly generate rotation angles $\bm{a} \sim U(0, \bm{A})$ for all the dimensions. After that, we uniformly sample scale factors $\bm{c} \sim U(\bm{C}_{min}, \bm{C}_{max})$. If the element in $\bm{c}$ is smaller than one, it shrinks the moving image. Otherwise, it enlarges the moving image. We then uniformly sample a translation factor $\bm{l} \sim U(-\bm{L}, \bm{L})$. To model the elastic distortion, we firstly randomly generate coordinate offset from Gaussian distribution with standard deviation $\gamma \sim U(0, \Gamma)$. We further apply a multidimensional Gaussian filter with standard deviation $\sigma \sim U(\bm{\Sigma}_{min}, \bm{\Sigma}_{max})$ to make the generated coordinate offset smooth and realistic.

During training, we generate the registration field ground truth $\bm{F}_g^0$ and fixed image $\bm{I}^0$ on the fly for each batch
\begin{equation}\label{eq:reg_sim_im}
\begin{aligned}
    \bm{F}_g^0 &= \mathcal{S}(\bm{a}, \bm{c}, \bm{l}, \gamma, \sigma| \bm{M}),\\
    \bm{I}^0 &= \mathcal{T}(\bm{M}, \bm{F}_g^0).
\end{aligned}
\end{equation}
For the registration-based segmentation, the segmentation ground truth $\bm{S}_g^0$ of fixed image $\bm{I}^0$ can be generated by
\begin{equation}\label{eq:reg_sim_label}
    \bm{S}_g^0 = \mathcal{T}(\bm{S}^M, \bm{F}_g^0),
\end{equation}
where $\bm{S}^M$ is the segmentation ground truth of the moving/atlas image $\bm{M}$ in the training set.
\subsection{Neural Registration}\label{sec:nr}
After we obtain the fixed image $\bm{I}^0$ with moving image $\bm{M}$, we design the neural registration to estimate the registration field $\bm{F}^0$. We employ a neural network $\mathcal{N}$ to parameterize the registration field
\begin{equation}\label{eq:nr}
    \bm{F}^0 = \mathcal{N}(\bm{M}, \bm{I}^0; \bm{\theta}),
\end{equation}
where $\bm{\theta}$ is the parameters in the neural network.

Different from unsupervised transformation estimation, we design a registration field supervised loss $\mathcal{L}_F$ based on generated registration field ground truth $\bm{F}_g^0$ from registration simulator
\begin{equation}\label{eq:l_f}
    \mathcal{L}_F(\bm{F}^0, \bm{F}_g^0; \bm{\theta}) = \frac{1}{|\Omega|}\sum_{\bm{p} \in \Omega}{|| \bm{F}^0(\bm{p}) - \bm{F}_g^0(\bm{p})||_{L_2}}, 
\end{equation}
where $\bm{p}$ is the pixel position in the image coordinate space $\Omega$. The field supervised loss in Eq.~\ref{eq:l_f} is the endpoint error (EPE) which is an accurate loss for image matching measuring the alignment of two displacement fields~\cite{zhu2017guided}. The registration field ground truth is smooth and the model can learn the smoothness of estimated registration field from registration simulator $\mathcal{S}$. Unlike the bending energy loss used in other unsupervised transformation estimation~\cite{balakrishnan2018unsupervised}, the registration field supervised loss is accurate and the loss can steadily decrease with the decrease of data similarity loss experimentally in Fig.~\ref{fig:reg_loss}. 

However, the model might heavily rely on the quality and diversity of the simulated registration field. To improve the model's generalization ability, we further employ data similarity loss as an auxiliary loss inspired by the unsupervised transformation estimation. To train the network in an end-to-end manner, a spatial transformer network $\mathcal{T}$ is used to obtain the reconstructed image $\bm{I}_R^0$ by warping the moving image $\bm{M}$ with the estimated registration field $\bm{F}^0$~\cite{jaderberg2015spatial}
\begin{equation}\label{eq:warp_stn}
    \bm{I}_R^0 = \mathcal{T}(\bm{M}, \bm{F}^0).
\end{equation}
We use the negative normalized local cross-correlation which is robust to evaluate similarity between MRI images
\begin{equation}\label{eq:l_sim}
\begin{aligned}
&\mathcal{L}_{sim}(\bm{I}^0, \bm{I_R}^0; \bm{\theta}) = \\ &-\frac{1}{|\Omega|}\sum_{\bm{p} \in \Omega} \frac{ \big( \sum_{\bm{p}_i} {(\bm{I}^0(\bm{p}_i) - \overline{\bm{I}^0(\bm{p})} ) (\bm{I}_R^0(\bm{p}_i) - \overline{\bm{I}_R^0(\bm{p})} )} \big)^2}{\sum_{\bm{p}_i} (\bm{I}^0(\bm{p}_i) - \overline{\bm{I}^0(\bm{p})} )^2   \sum_{\bm{p}_i} (\bm{I}_R^0(\bm{p}_i) - \overline{\bm{I}_R^0(\bm{p})} )^2 }, 
\end{aligned}
\end{equation}
where $\bm{p}_i$ is the pixel position within a window around $\bm{p}$, and $\overline{\bm{I}^0(\bm{p})}$ and $\overline{\bm{I}_R^0(\bm{p})}$ are local means within the window around pixel position $\bm{p}_i$ in $\bm{I}^0$ and $\bm{I}_R^0$ respectively.

For NeurReg, we employ the hybrid loss between data similarity loss and registration field supervised loss to train the neural network $\mathcal{N}$
\begin{equation}\label{eq:loss_reg}
\begin{aligned}
    \mathcal{L}_{reg}(\bm{F}^0, \bm{F}_g^0, \bm{I}^0, \bm{I}_R^0; \bm{\theta}) =& \mathcal{L}_F(\bm{F}^0, \bm{F}_g^0; \bm{\theta}) \\ &+ \lambda \mathcal{L}_{sim}(\bm{I}^0, \bm{I}_R^0; \bm{\theta}),
\end{aligned}
\end{equation}
where $\lambda$ is the hyper-parameter to balance the two losses. In the inference, given a moving/atlas image $\bm{M}$ and a fixed image $\bm{I}$, the registration field $\bm{F}$ can be estimated instantly by Eq.~\ref{eq:nr} and reconstructed image $\bm{I}_R$ can be further calculated from Eq.~\ref{eq:warp_stn}.
\subsection{Registration-Based Segmentation}
\label{sec:seg}
The estimated registration field can be applied to transforming segmentation mask for image segmentation purpose. From section~\ref{sec:nr}, we obtain the estimated registration field $\bm{F}$ from Eq.~\ref{eq:nr} given a moving/atlas image $\bm{M}$ and a test image as the fixed image $\bm{I}$. With the segmentation ground truth $\bm{S}^M$ of moving image, we can further obtain the predicted segmentation mask $\bm{S}$ through warping $\bm{S}^M$ with nearest neighbor re-sampling.

More importantly, if ground truths of the segmentation are available during training, we can utilize them to further tune the registration network. This joint learning of segmentation and registration tasks can be beneficial for both because the tasks are highly correlated~\cite{ruder2017overview}.

In registration-based segmentation by multi-task learning (MTL), we introduce a dual registration scheme to fully exploit the power of registration simulator in the proposed NeurReg as illustrated in Fig.~\ref{fig:framework}. We expect the moving/atlas image $\bm{M}$ to be aligned with images $\bm{I}^1$ and $\bm{I}^0$ from both the dataset and registration simulator $\mathcal{S}$. In the segmentation scenario, the registration simulator $\mathcal{S}$ acts as a data augmentation which is crucial to medical image segmentation because medical image dataset is typically small and dense annotation of segmentation is expensive. 

The other registration in dual registration for moving/atlas image $\bm{M}$ and a random image $\bm{I}^1$ from dataset can be obtained
\begin{equation}\label{eq:dual_reg}
\begin{aligned}
    \bm{F}^1 &= \mathcal{N}(\bm{M}, \bm{I}^1; \bm{\theta}), \\
    \bm{I}_R^1 &= \mathcal{T}(\bm{M}, \bm{F}^1).
\end{aligned}
\end{equation}
Given the segmentation ground truth $\bm{S}^M$ of a moving/atlas image $\bm{M}$, we can obtain the segmentation ground truth $\bm{S}_g^0$ from Eq.~\ref{eq:reg_sim_label}. If the segmentation ground truth $\bm{S}_g^1$ is available, we can further design a segmentation loss based on dual registration into the framework with Tversky loss~\cite{salehi2017tversky,zhu2019anatomynet}
\begin{equation}\label{eq:loss_seg}
\begin{aligned}
    \mathcal{L}&_{seg}(\bm{S}^0, \bm{S}_g^0, \bm{S}^1, \bm{S}_g^1) = \mathcal{D}(\bm{S}^0, \bm{S}_g^0) + \mathcal{D}(\bm{S}^1, \bm{S}_g^1), \\
    \mathcal{D}&(\bm{S}^0, \bm{S}_g^0) = - \frac{1}{C} \sum_{c=0}^{C-1} \frac{\sum_{\bm{p}}{2 \bm{S}^0(\bm{p}) \bm{S}_g^0(\bm{p})}}{ \sum_{\bm{p}}{\bm{S}^0(\bm{p}) + \bm{S}_g^0(\bm{p})}},
\end{aligned} 
\end{equation}
where $\bm{S}_g^0$ and $\bm{S}_g^1$ are one hot form of segmentation ground truth and $\bm{S}^0$ and $\bm{S}^1$ are continuous values with interpolation order one in the spatial transformer $\mathcal{T}$, $C$ is the number of classes. The loss function $\mathcal{L}_{reg}^{MTL}$ in the MTL turns to 
\begin{equation}\label{eq:loss_reg_mtl}
\begin{aligned}
    \mathcal{L}_{reg}^{MTL}&(\bm{F}^0, \bm{F}_g^0, \bm{I}^0, \bm{I}_R^0, \bm{I}^1, \bm{I}_R^1; \bm{\theta}) = \mathcal{L}_F(\bm{F}^0, \bm{F}_g^0; \bm{\theta}) \\ &+ \lambda \big( \mathcal{L}_{sim}(\bm{I}^0, \bm{I}_R^0; \bm{\theta}) + \mathcal{L}_{sim}(\bm{I}^1, \bm{I}_R^1; \bm{\theta}) \big) \\ &+ \beta  \mathcal{L}_{seg}(\bm{S}^0, \bm{S}_g^0, \bm{S}^1, \bm{S}_g^1),
\end{aligned}
\end{equation}
where $\beta$ controls the weight of segmentation loss. At inference time, we obtain the final segmentation prediction by taking $\mathrm{argmax}$ along the class dimension.

\begin{figure}[t]
\begin{center}
   \includegraphics[width=\linewidth]{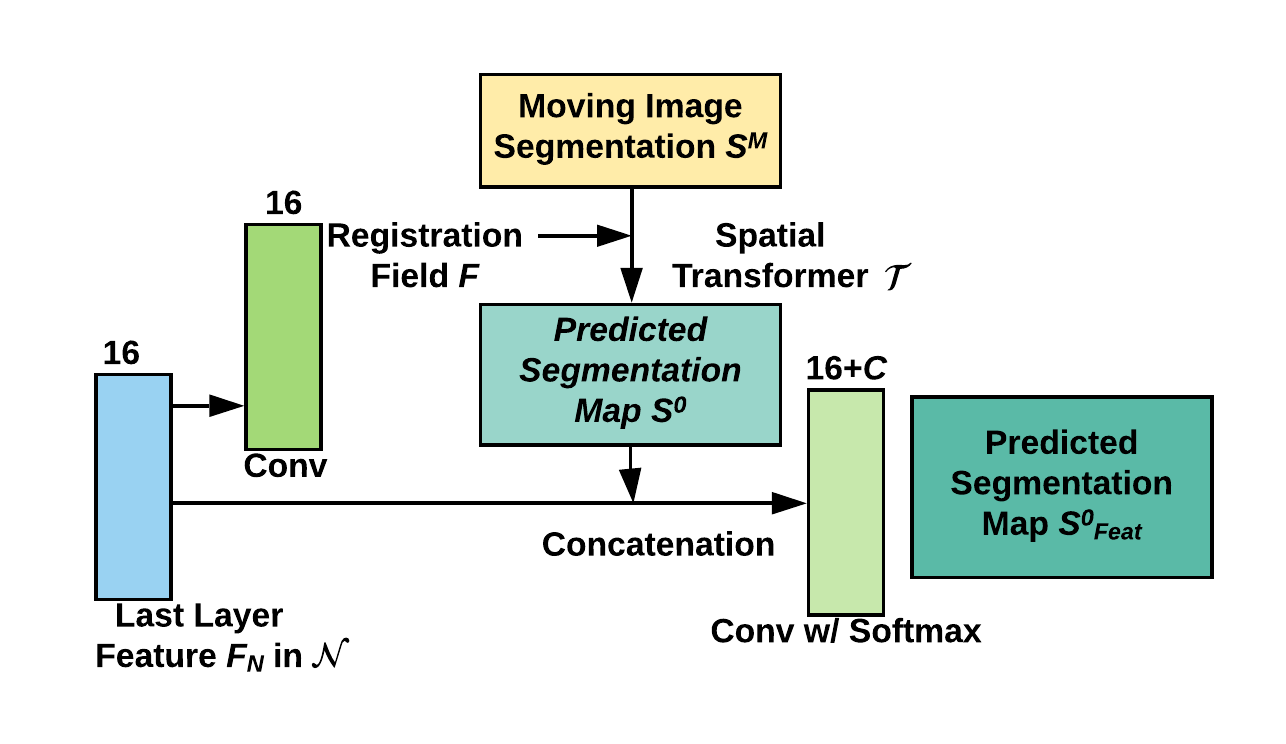}
\end{center}
   \caption{A residual block designed to boost the segmentation.}
\label{fig:boost_segm}
\end{figure}

Inspired by residual network and boosting concept~\cite{he2016deep,friedman2001greedy}, we further extend the framework by introducing an extra convolutional layer with predicted segmentation $\bm{S}^0$ and the last layer feature $\bm{F}_N$ from neural network $\mathcal{N}$ as input and softmax activation function to further improve the segmentation as illustrated in Fig.~\ref{fig:boost_segm}
\begin{equation}\label{eq:boost_segm}
    \bm{S}_{Feat}^0 = \mathrm{Softmax}(\mathrm{Conv}([\bm{F}_N, \bm{S}^0])),
\end{equation}
where $\bm{S}_{Feat}^0$ is the segmentation prediction using aligned segmentation prediction $\bm{S}^0$ as feature and $[\cdot, \cdot]$ is the concatenation along channel dimension. The segmentation loss based on Tversky loss in Eq.~\ref{eq:loss_seg} can be easily adapted in the feature based segmentation. 

\section{Experiments}
We conduct experiments on the Hippocampus and Prostate MRI datasets from medical segmentation decathlon to fully validate the proposed NeurReg~\cite{simpson2019large}.
\subsection{Datasets and Experimental Settings}
On the Hippocampus dataset, we randomly split the dataset into 208 training images and 52 test images. There are two foreground categories in the segmentation, hippocampus head and hippocampus body. Because the image size is within $48 \times 64 \times 48$ voxels, we use a $5 \times 5 \times 5$-voxel window in the similarity loss $\mathcal{L}_{sim}$ in Eq.~\ref{eq:l_sim}.

On the Prostate dataset, we randomly split the dataset into 25 training images and seven test images. The Prostate dataset consists of T2 weighted and apparent diffusion coefficient (ADC) MRI scans. Because of the low signal to noise ratio in ADC scans, we only use the T2 weighted channel. There are two foreground categories, prostate peripheral zone and prostate central gland. Because the image size is around $96 \times 240 \times 240$ which is larger than the Hippocampus dataset, we use a $9 \times 9 \times 9$-voxel window in Eq.~\ref{eq:l_sim}.

We re-sample the MR images to $1\times 1 \times 1$ $\mathrm{mm}^3$ spacing. To reduce the discrepancy of the intensity distributions of the MR images, we calculate the mean and standard deviation of each volume and clip each volume up to six standard deviations. Finally, we linearly transform each 3D image into range $[0, 1]$. 

The hyper-parameters in the registration simulator $\mathcal{S}$ are empirically set as $\bm{A}=(\frac{1}{6}\pi, \frac{1}{6}\pi, \frac{1}{6}\pi)$, $\bm{C}_{min}=(0.75, 0.75, 0.75)$, $\bm{C}_{max}=(1.25, 1.25, 1.25)$, $\bm{L}=(0.02, 0.02, 0.02)$, $\Gamma=1000$, $\Sigma_{min}=10$, $\Sigma_{min}=13$ based on data distribution. We use 3D U-Net type network as $\mathcal{N}$~\cite{ronneberger2015u,cciccek20163d}. There are four layers in the encoder with numbers of channels $(16, 32, 32, 32)$ with stride as two in each layer. After that we use two convolutional layers with number of channels $32$ and $32$. For the decoder, we use four blocks of up-sampling, concatenation and convolution with numbers of channels $(32, 32, 32, 16)$. Finally, we use a convolutional layer with number of channels of $16$. LeakyReLU is used with slop of 0.2 in each convolutional layer in the encoder and decoder~\cite{maas2013rectifier}. We use Adam optimizer with learning rate $10^{-4}$~\cite{kingma2014adam}, and numbers of epochs of 1,500 and 2,000 for the Hippocampus and Prostate datasets respectively. Because of small prostate dataset, large image size and slow training, we initialize the model by pretrained model on Hippocampus dataset. Because the registration field supervised loss in Eq.~\ref{eq:l_f} is relative large, we set $\lambda$ and $\beta$ as 10 to balance the three losses. Most of the hyper-parameter settings are the same as VoxelMorph for fair comparisons~\cite{balakrishnan2018unsupervised}. And we further use the recommended hyper-parameters for ANTs and NiftyReg in~\cite{balakrishnan2018unsupervised}, because the field of view in the hippopotamus and prostate images is already roughly aligned during image acquisition. We use three scales with 200 iterations each, B-Spline SyN step size of 0.25, updated field mesh size of five for ANTs. For NiftyReg, we use three scales, the same local negative cross correlation objective function with control point grid spacing of five voxels and 500 iterations. 

\subsection{Registration Performance Comparisons}
To construct a registration test dataset which can be used to quantitatively compare different registrations, we randomly generate two registration fields for each test image in the test dataset. We use five evaluation metrics in the registration, average time in seconds for inference on one pair of images, average registration field endpoint error (EPE) which is the same as Eq.~\ref{eq:l_f}, average mean square error over reconstructed image and fixed image (MSE), average normalized local cross-correlation with neighbor volume size $5\times 5 \times 5$ the same as negative value in Eq.~\ref{eq:l_sim} (NLCC), and average mutual information with 100 bins (MI). The comprehensive comparisons over ANTs~\cite{avants2009advanced}, NiftyReg~\cite{modat2010fast} and VoxelMorph~\cite{balakrishnan2018unsupervised} are listed in the table~\ref{tab:reg_hippo} and~\ref{tab:reg_prost} for the Hippocampus and Prostate datasets respectively.
{\small \begin{table*}
\begin{center}
\begin{tabular}{|l|c|c|c|c|c|}
\hline
Method & Time (s) & EPE (mm) & MSE & NLCC & MI \\
\hline\hline
ANTs &289.05$\pm$135.71 &4.267$\pm$1.809 &0.008$\pm$0.006 &0.624$\pm$0.115 &0.934$\pm$ 0.276\\
NiftyReg & 991.48$\pm$420.90& 4.113$\pm$1.450 &0.002$\pm$0.002& 0.795$\pm$0.058 & 1.484$\pm$ 0.195 \\
VoxelMorph &0.03$\pm$ 0.18 &6.222$\pm$ 1.573 &0.004$\pm$ 0.002 &0.723$\pm$ 0.047 &1.006$\pm$ 0.134 \\
VoxelMorph (MTL) & 0.03$\pm$0.14 & 6.236$\pm$1.579 & 0.007$\pm$0.004 & 0.598$\pm$0.050 & 0.775$\pm$0.120 \\
VoxelMorph (Feat.) &0.04$\pm$0.16 &6.242$\pm$1.573 &0.005$\pm$0.003 &0.687$\pm$0.050 &0.911$\pm$0.133 \\
\hline
Ours (NeurReg) &0.03$\pm$0.15 &$0.957\pm 0.312$ &${\bm{0.001\pm 0.153}}$ &${\bm{0.808\pm 0.069}}$ &${\bm{1.520\pm 0.191}}$ \\
Ours w/o $\mathcal{L}_{sim}$ & 0.03$\pm$0.18&${\bm{0.950\pm 0.330}}$ &$0.002\pm 0.002$ &0.762$\pm$0.087 &$1.386\pm 0.219$ \\
Ours (MTL) &0.06$\pm$0.19 &1.277$\pm$0.382 &0.002$\pm$0.001 &0.749$\pm$0.070 &1.337$\pm$0.164 \\
Ours (Feat.) &0.05$\pm$0.20 &1.146$\pm$0.339 &0.002$\pm$0.001 &0.782$\pm$0.068 &1.410$\pm$0.176 \\
\hline
\end{tabular}
\end{center}
\caption{Registration comparisons on the Hippocampus dataset. NeurReg is the best based on all the metrics. Best scores are in bold face.}
\label{tab:reg_hippo}
\end{table*}}

{\small  \begin{table*}
\begin{center}
\begin{tabular}{|l|c|c|c|c|c|}
\hline
Method & Time (s) & EPE (mm) & MSE & NLCC & MI \\
\hline\hline
ANTs  &5851.84$\pm$2450.31 &24.371$\pm$8.535 &0.021$\pm$0.005 &0.304$\pm$0.074 &0.334$\pm$0.173 \\
NiftyReg  &2307.32$\pm$662.08 &25.556$\pm$7.595 &0.008$\pm$0.003 &0.407$\pm$0.096 &0.821$\pm$0.242 \\
VoxelMorph  & 0.92$\pm$1.71& 26.483$\pm$7.450&0.010$\pm$0.001 &${\bm{0.504\pm 0.033}}$ &0.472$\pm$0.104 \\
VoxelMorph (MTL) &1.07$\pm$1.60 &26.480$\pm$7.432 &0.012$\pm$0.002 &0.423$\pm$0.029 &0.376$\pm$0.089 \\
VoxelMorph (Feat.) &1.20$\pm$1.72 &26.489$\pm$7.441 &0.013$\pm$0.002 &0.433$\pm$0.027 &0.356$\pm$0.084 \\
\hline 
Ours (NeurReg) & 0.78$\pm$1.34&$5.228\pm 1.169$ &${\bm{0.006\pm 0.002}}$ &0.363$\pm$0.057 &${\bm{0.860\pm 0.231}}$ \\
Ours w/o $\mathcal{L}_{sim}$ & 0.76$\pm$1.33&${\bm{5.082\pm 1.173}}$ &$0.008\pm 0.002$ &0.280$\pm$0.052 &0.735$\pm 0.204$ \\ 
Ours (MTL) &1.90$\pm$3.12 &6.991$\pm$1.452 &0.008$\pm$0.002 &0.292$\pm$0.039 &0.700$\pm$0.198 \\
Ours (Feat.) &1.70$\pm$2.15 &6.532$\pm$1.687 &0.008$\pm$0.003 &0.321$\pm$0.045 &0.741$\pm$0.215 \\
\hline
\end{tabular}
\end{center}
\caption{Registration comparisons on the Prostate dataset.}
\label{tab:reg_prost}
\end{table*}}

From the table~\ref{tab:reg_hippo} and~\ref{tab:reg_prost}, our registration achieves substantial improvement on EPE which is one of the most accurate ways to evaluate the performance of registration in equation~\ref{eq:l_f}. Our registration obtains the best performance on the MSE which is a robust metric for data with the same distribution in the current scenario. The NeurReg yields the best MI on the two dataset and the best NLCC on the Hippocampus dataset. The hybrid loss obtains comparable EPE metric and does much better on the other metrics compared with model without similarity loss. The multi-task learning-based registration obtains the comparable registration performance as the base methods. The improvement over the three mostly used registration toolboxes on the two datasets with the five metrics confirms the robustness and good performance of the proposed NeurReg. The improvement probably is because the registration field guided learning in NeurReg leads to an optimal convergent point in the learning. 

\begin{figure}[t]
\begin{center}
        \begin{minipage}{0.19\linewidth}
			\includegraphics[width=\textwidth]{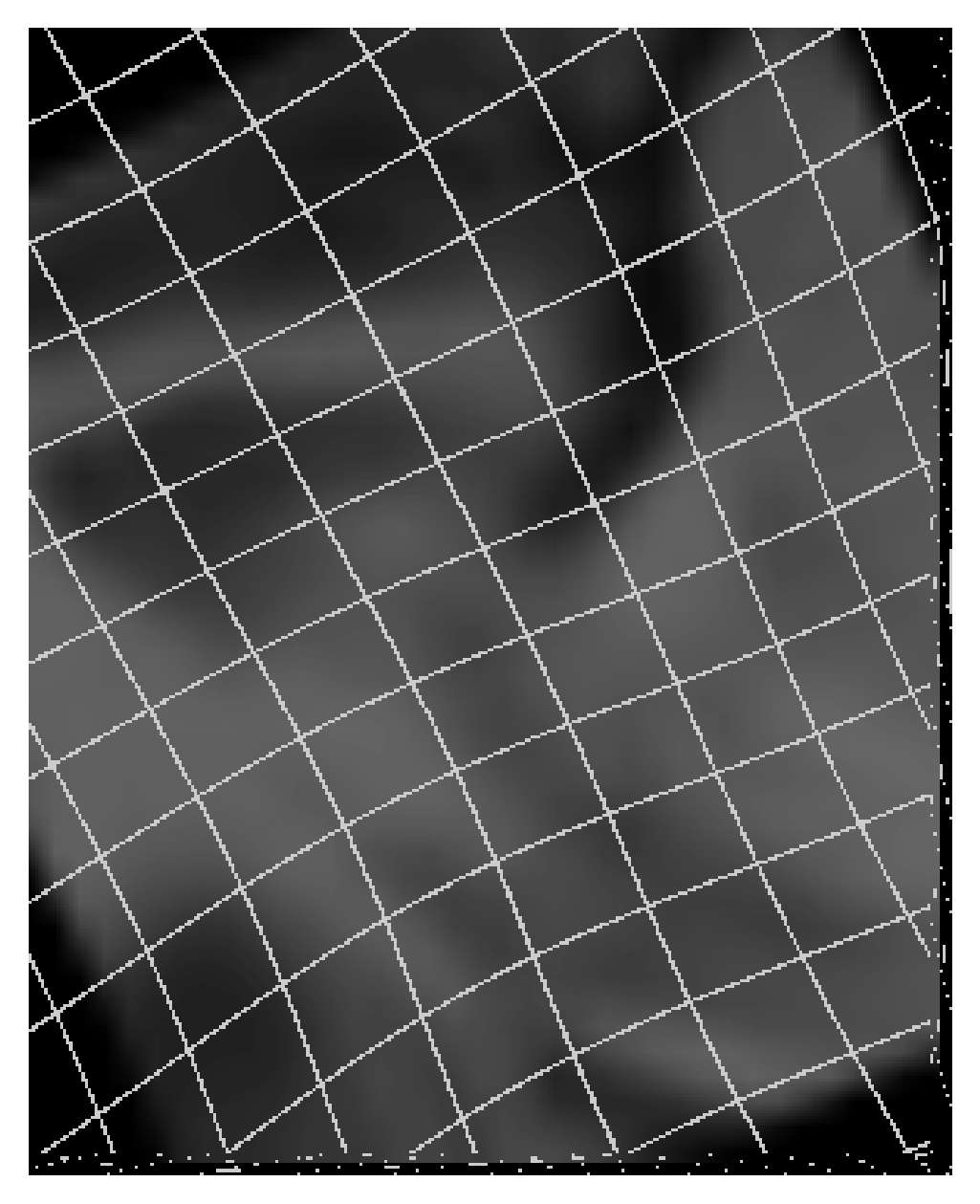}
		\end{minipage}
		\begin{minipage}{0.19\linewidth}
			\includegraphics[width=\textwidth]{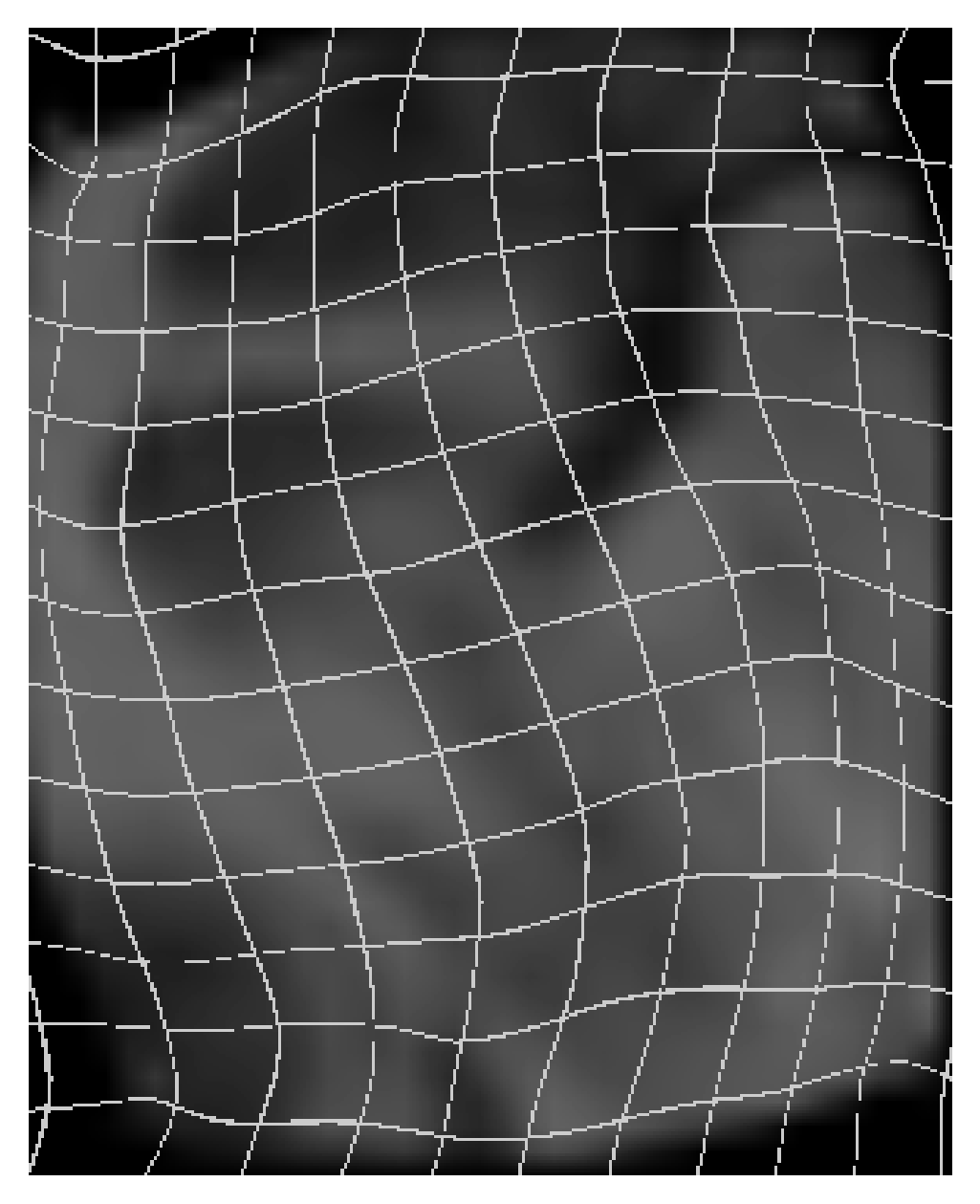}
		\end{minipage}
		\begin{minipage}{0.19\linewidth}
			\includegraphics[width=\textwidth]{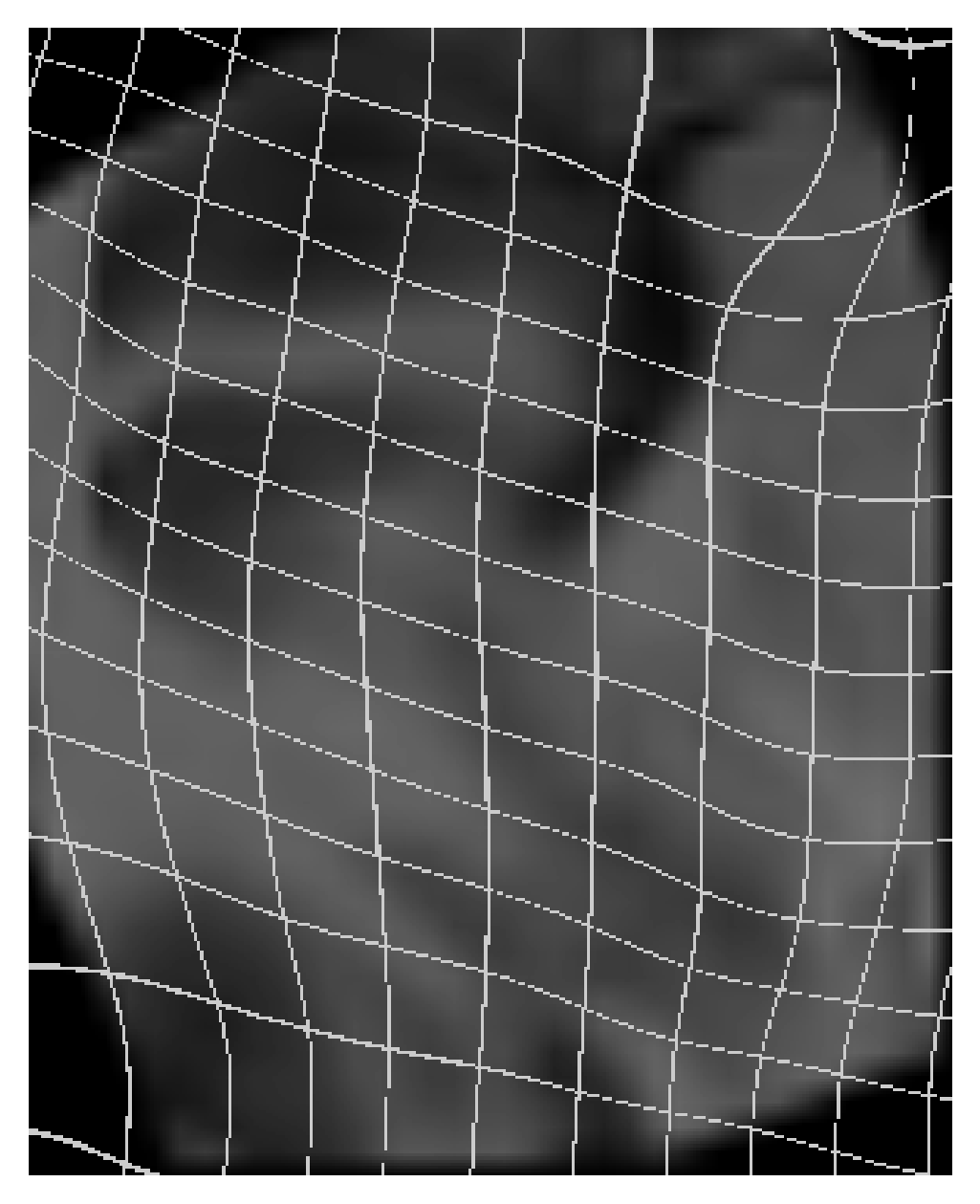}
		\end{minipage}
		\begin{minipage}{0.19\linewidth}
			\includegraphics[width=\textwidth]{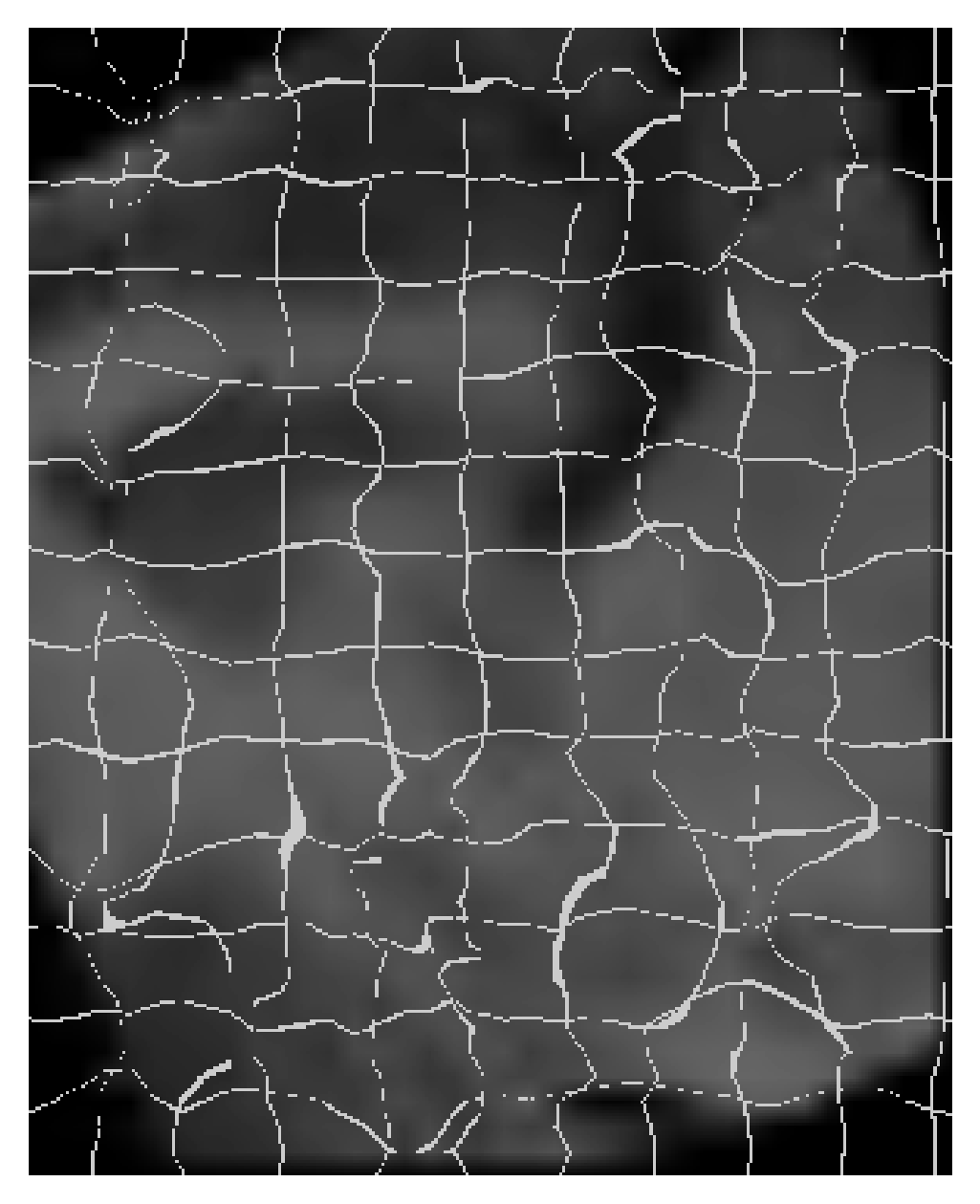}
		\end{minipage}
		\begin{minipage}{0.19\linewidth}
			\includegraphics[width=\textwidth]{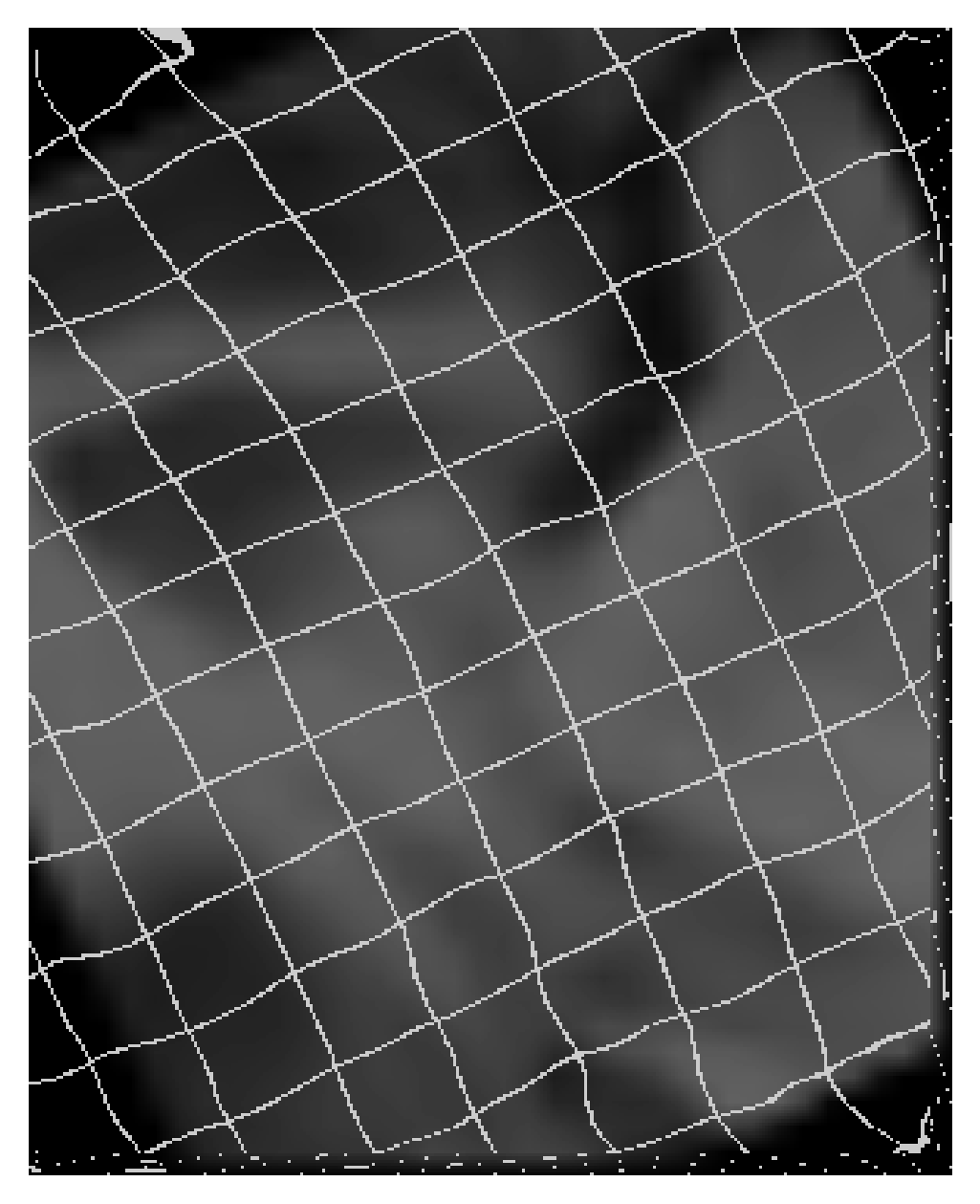}
		\end{minipage}\\
		
		\begin{minipage}{0.19\linewidth}
			\includegraphics[width=\textwidth]{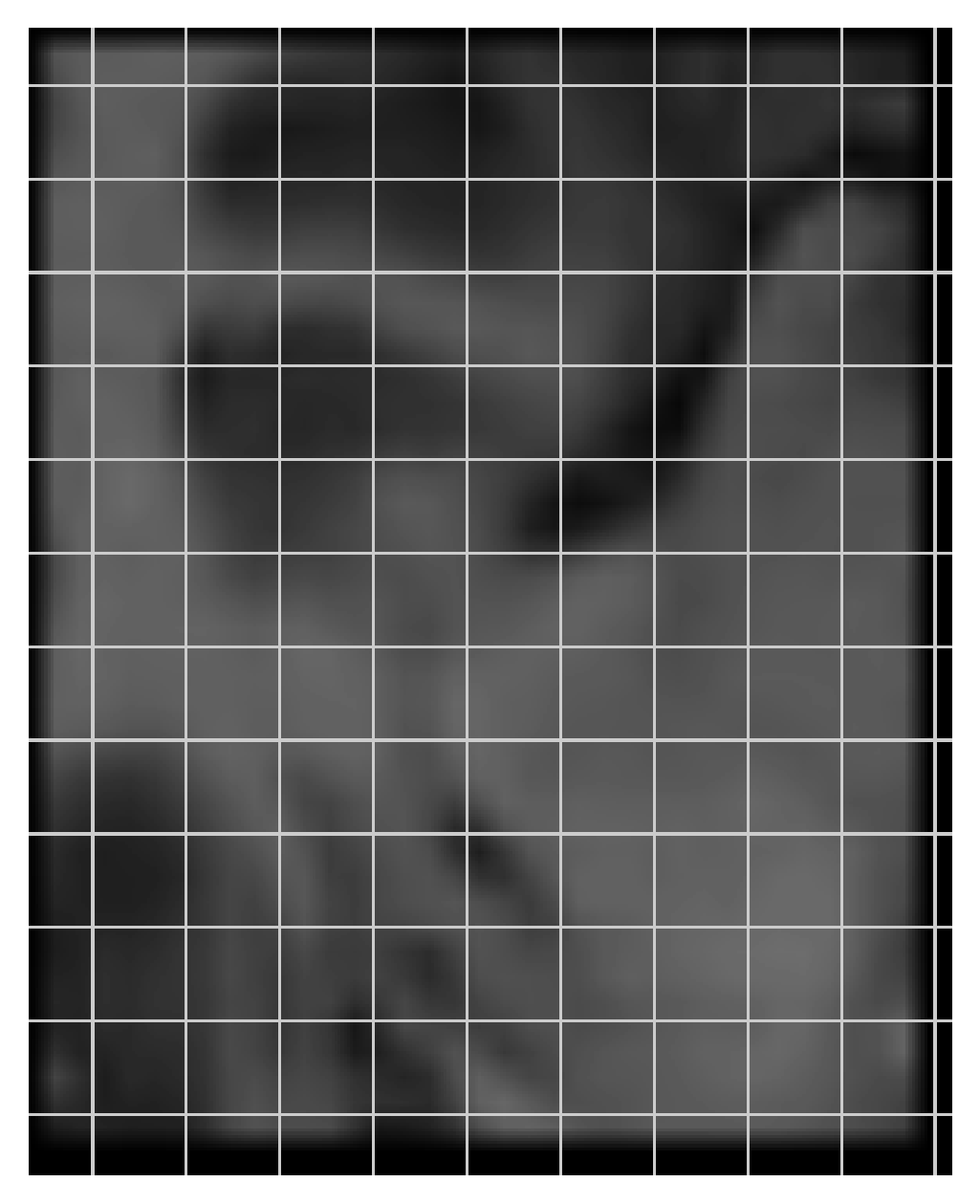}
		\end{minipage}
		\begin{minipage}{0.19\linewidth}
			\includegraphics[width=\textwidth]{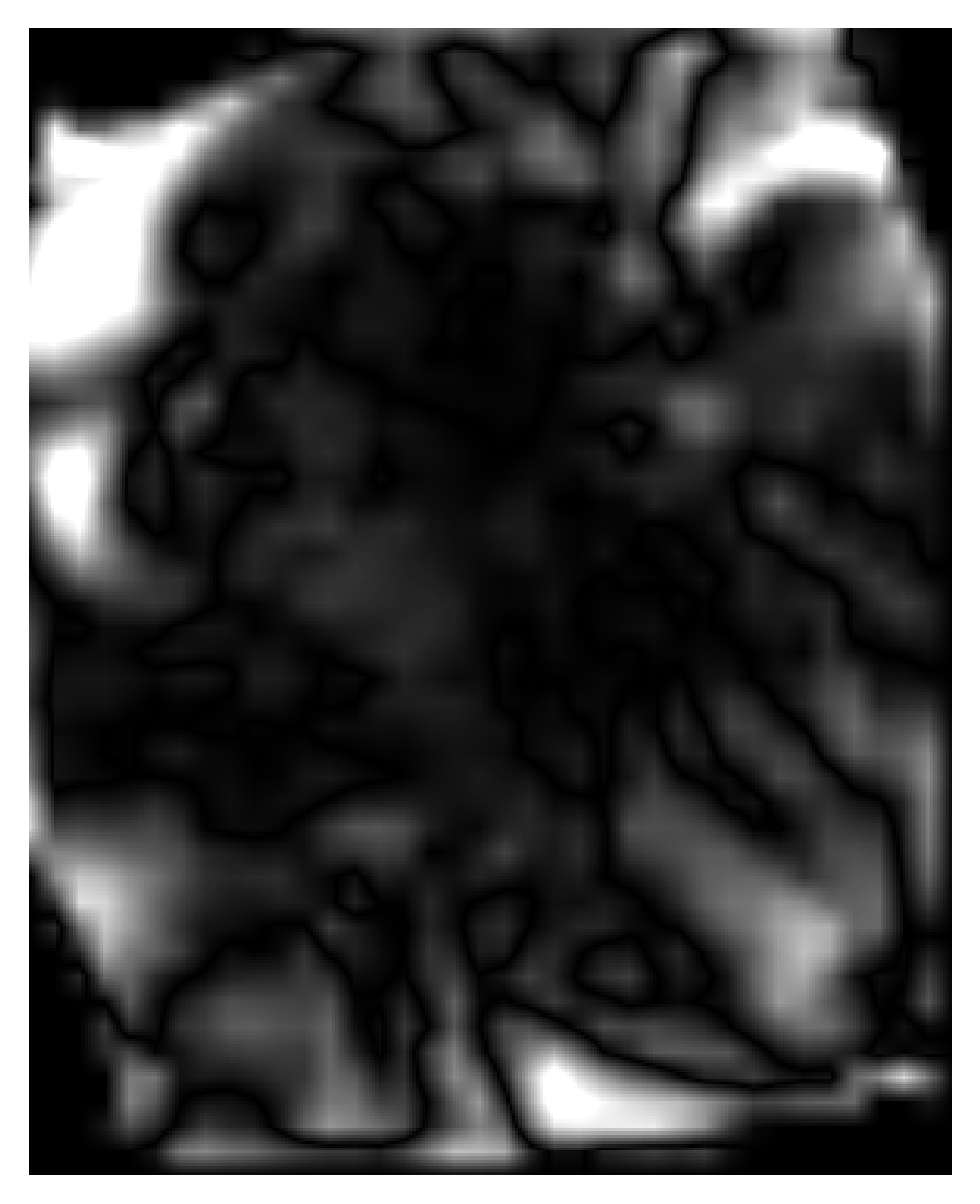}
		\end{minipage}
		\begin{minipage}{0.19\linewidth}
			\includegraphics[width=\textwidth]{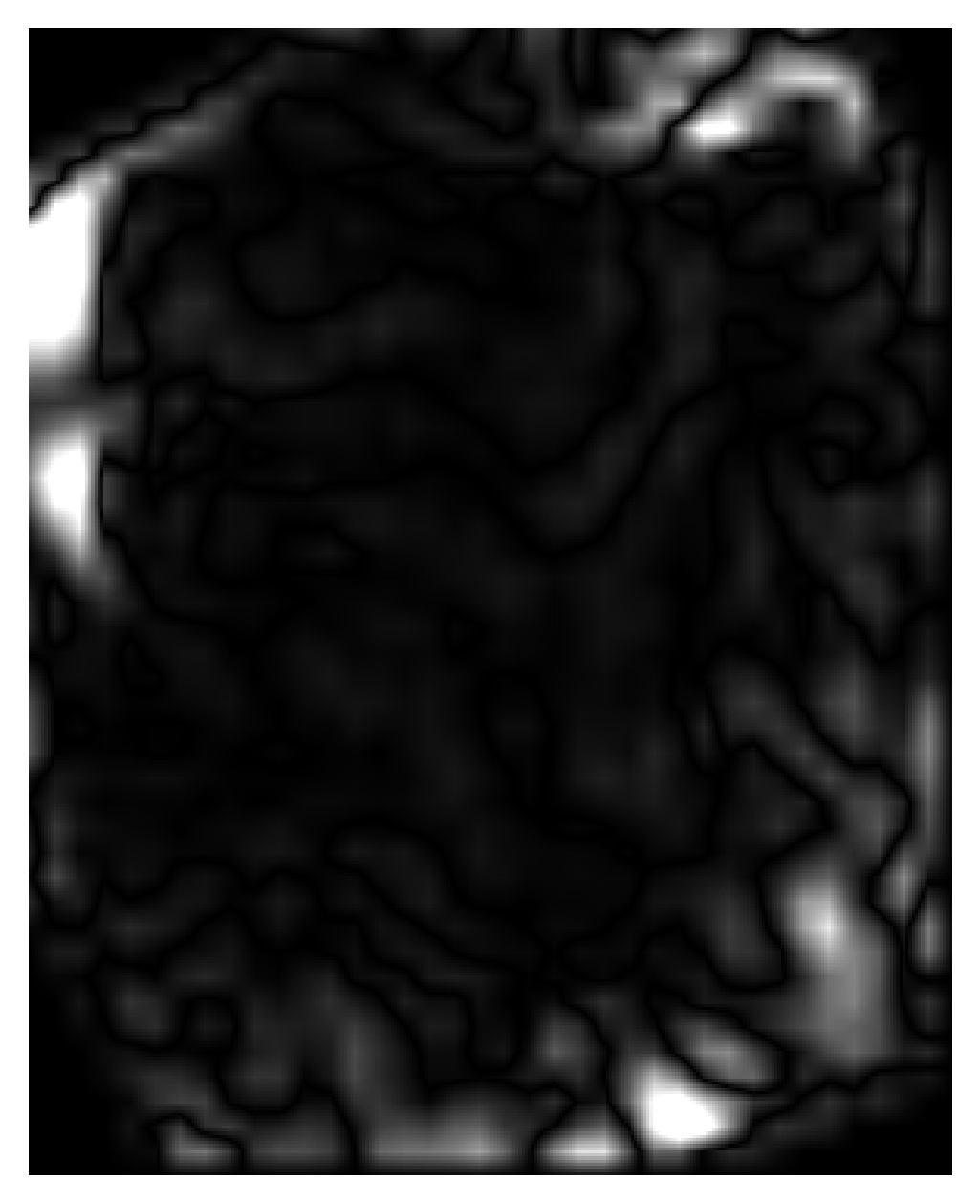}
		\end{minipage}
		\begin{minipage}{0.19\linewidth}
			\includegraphics[width=\textwidth]{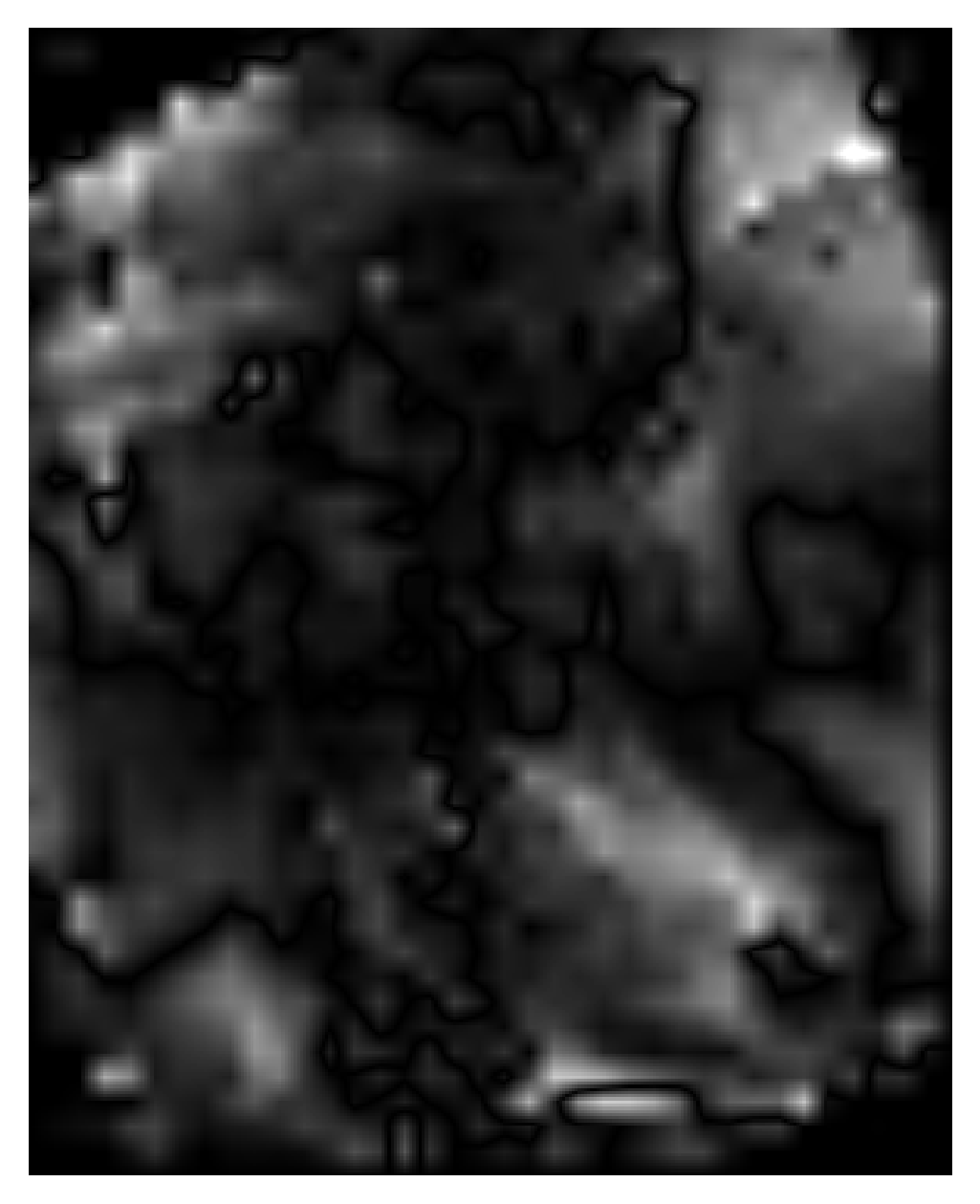}
		\end{minipage}
		\begin{minipage}{0.19\linewidth}
			\includegraphics[width=\textwidth]{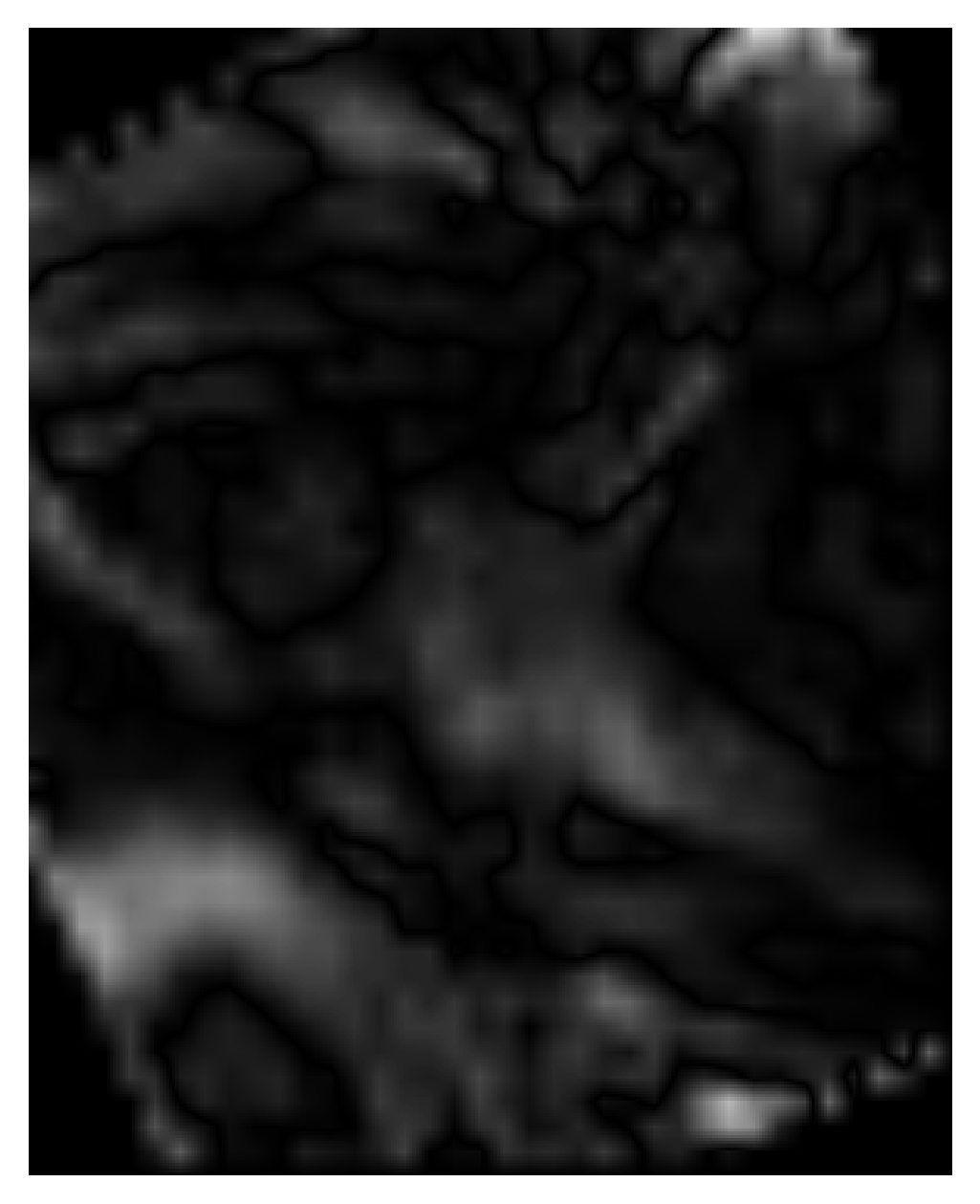}
		\end{minipage}\\
		
		\begin{minipage}{0.19\linewidth}
			\includegraphics[width=\textwidth]{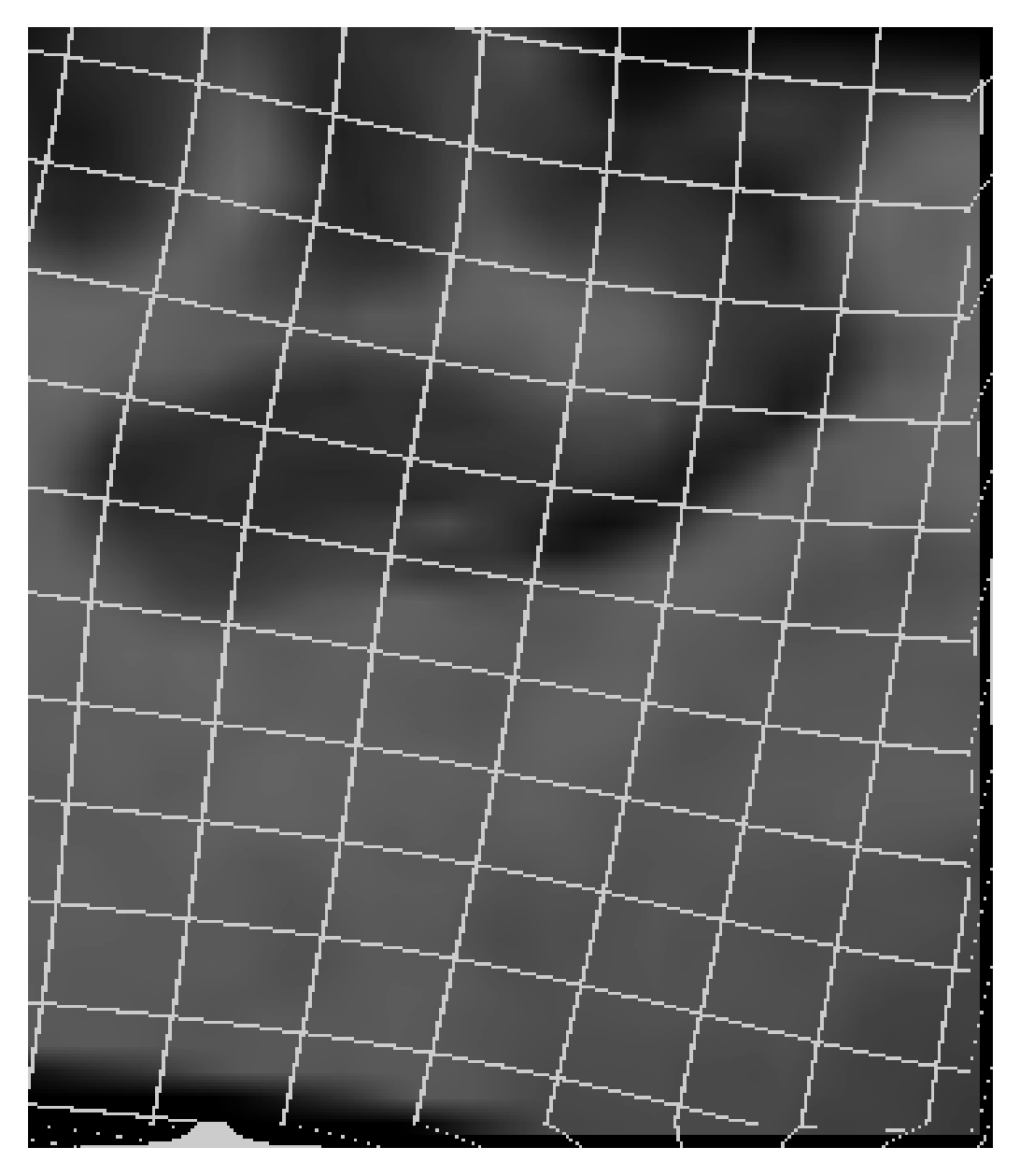}
		\end{minipage}
		\begin{minipage}{0.19\linewidth}
			\includegraphics[width=\textwidth]{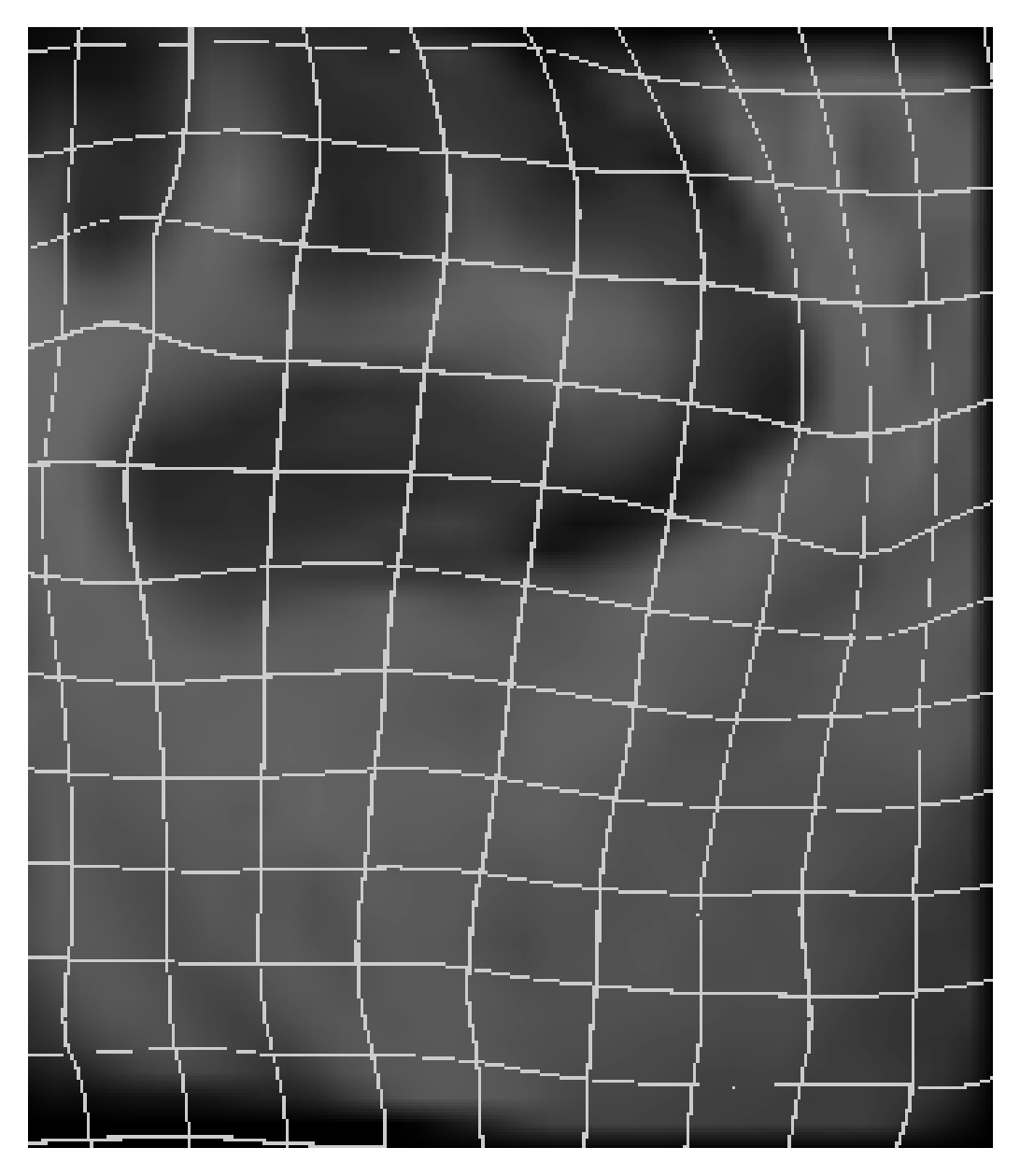}
		\end{minipage}
		\begin{minipage}{0.19\linewidth}
			\includegraphics[width=\textwidth]{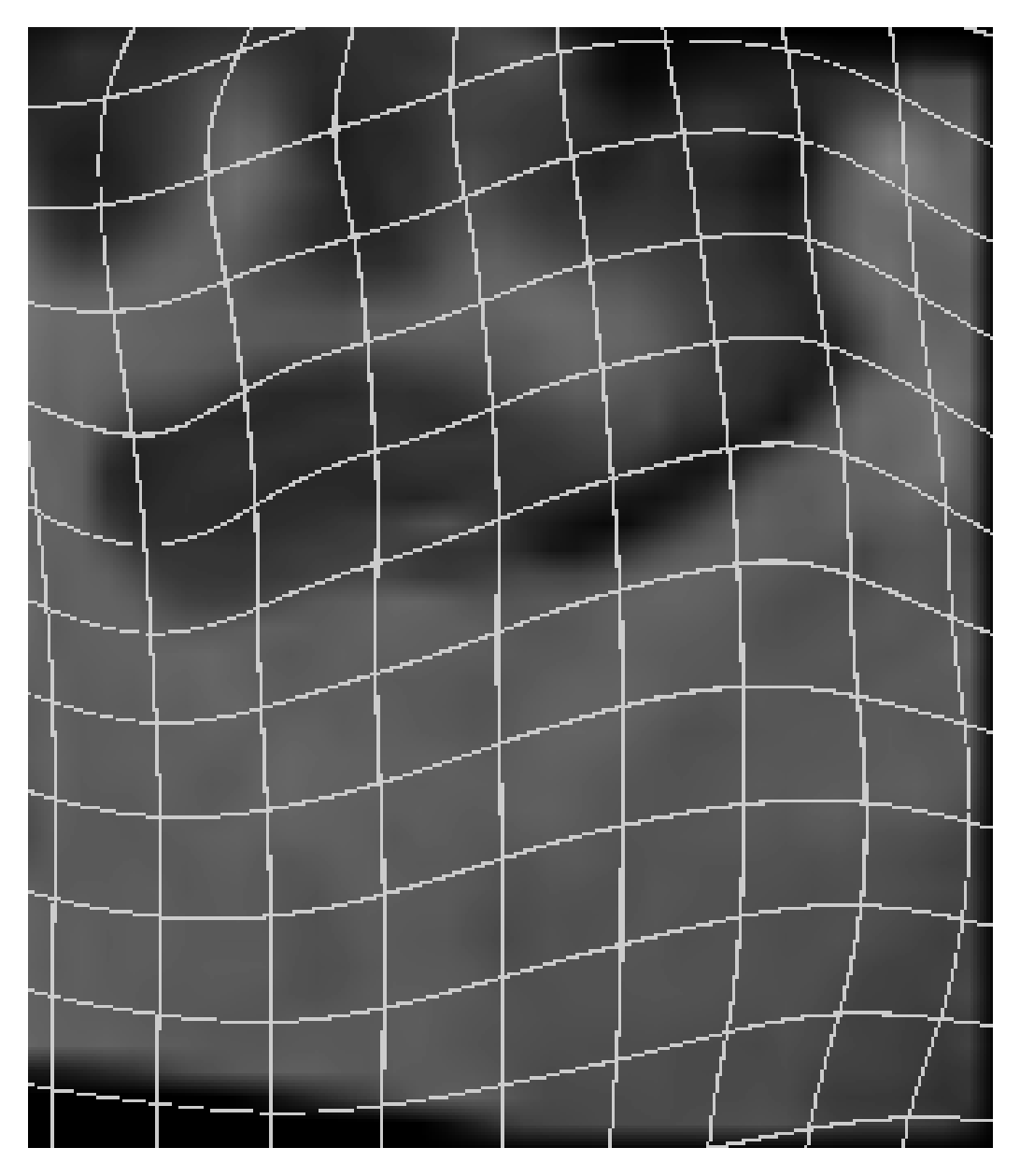}
		\end{minipage}
		\begin{minipage}{0.19\linewidth}
			\includegraphics[width=\textwidth]{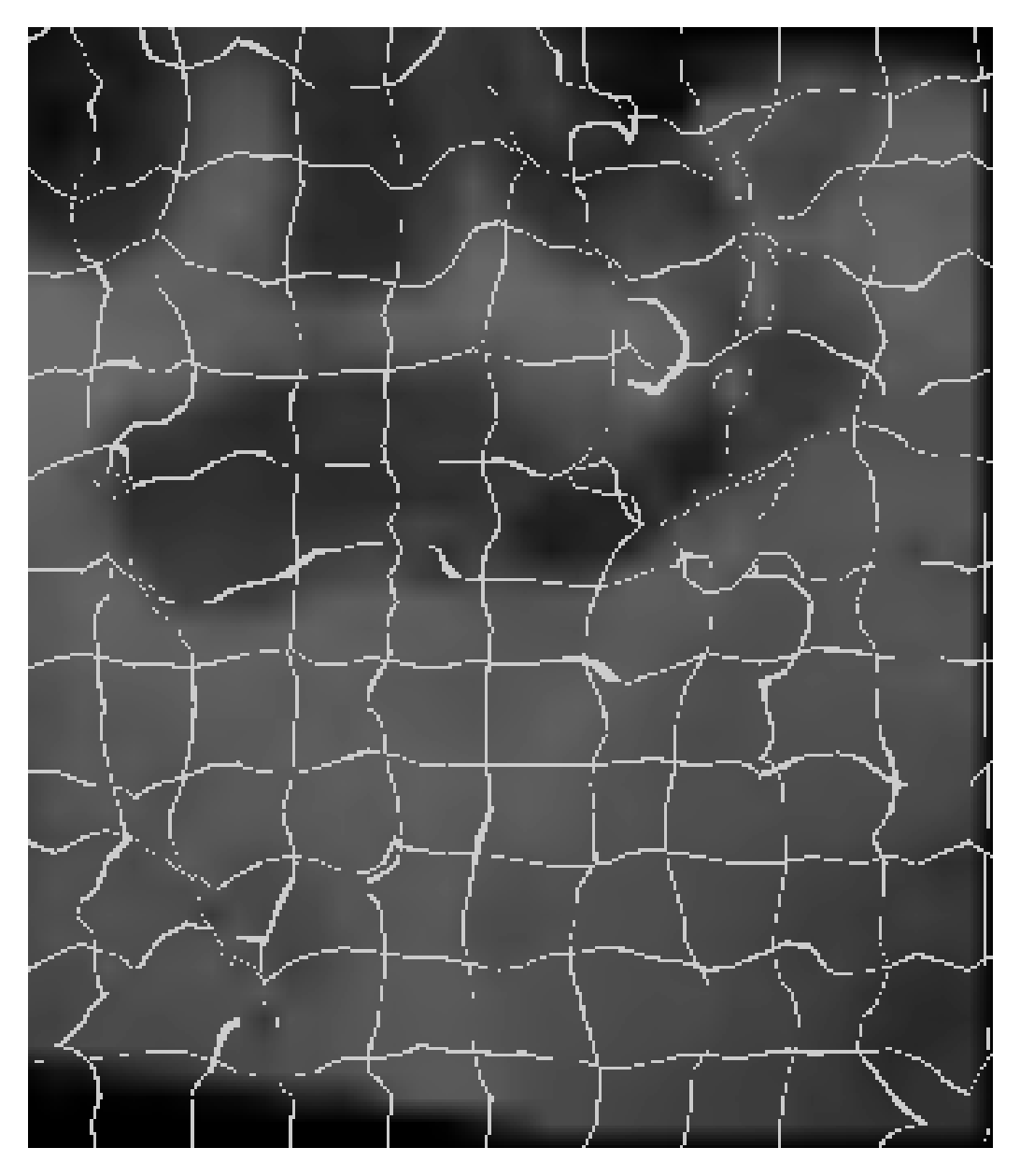}
		\end{minipage}
		\begin{minipage}{0.19\linewidth}
			\includegraphics[width=\textwidth]{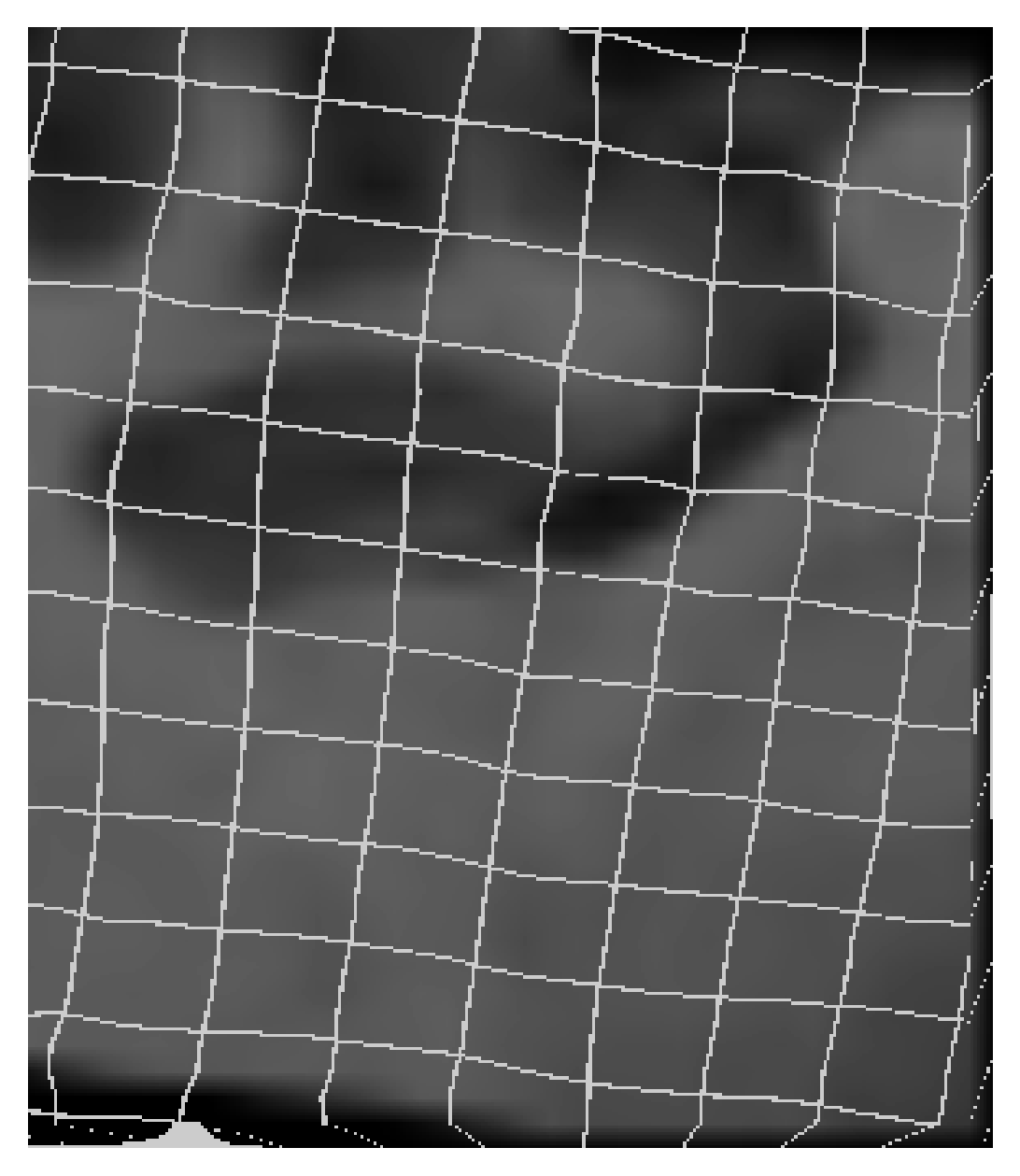}
		\end{minipage}\\
		
		\begin{minipage}{0.19\linewidth}
			\includegraphics[width=\textwidth]{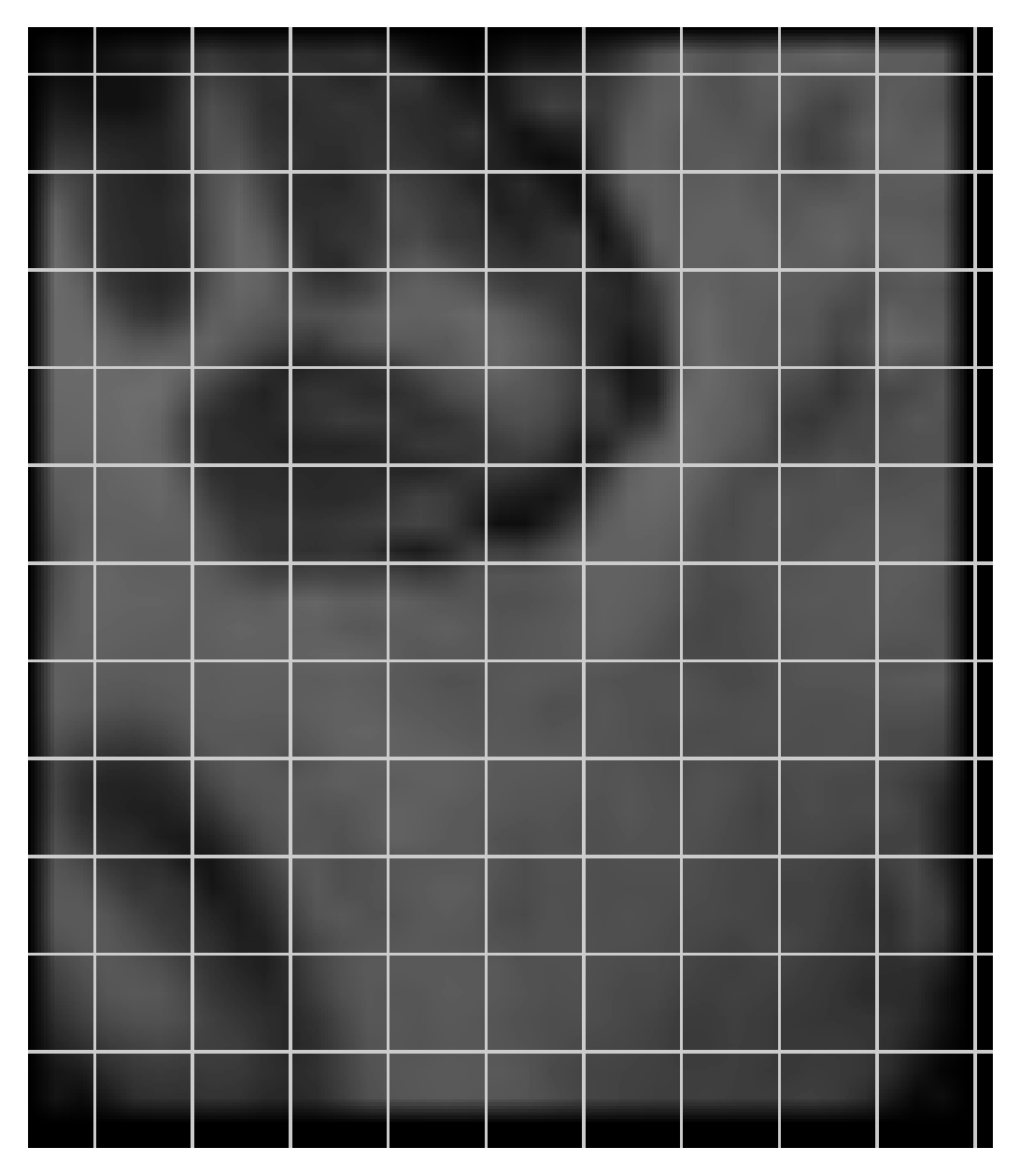}
		\end{minipage}
		\begin{minipage}{0.19\linewidth}
			\includegraphics[width=\textwidth]{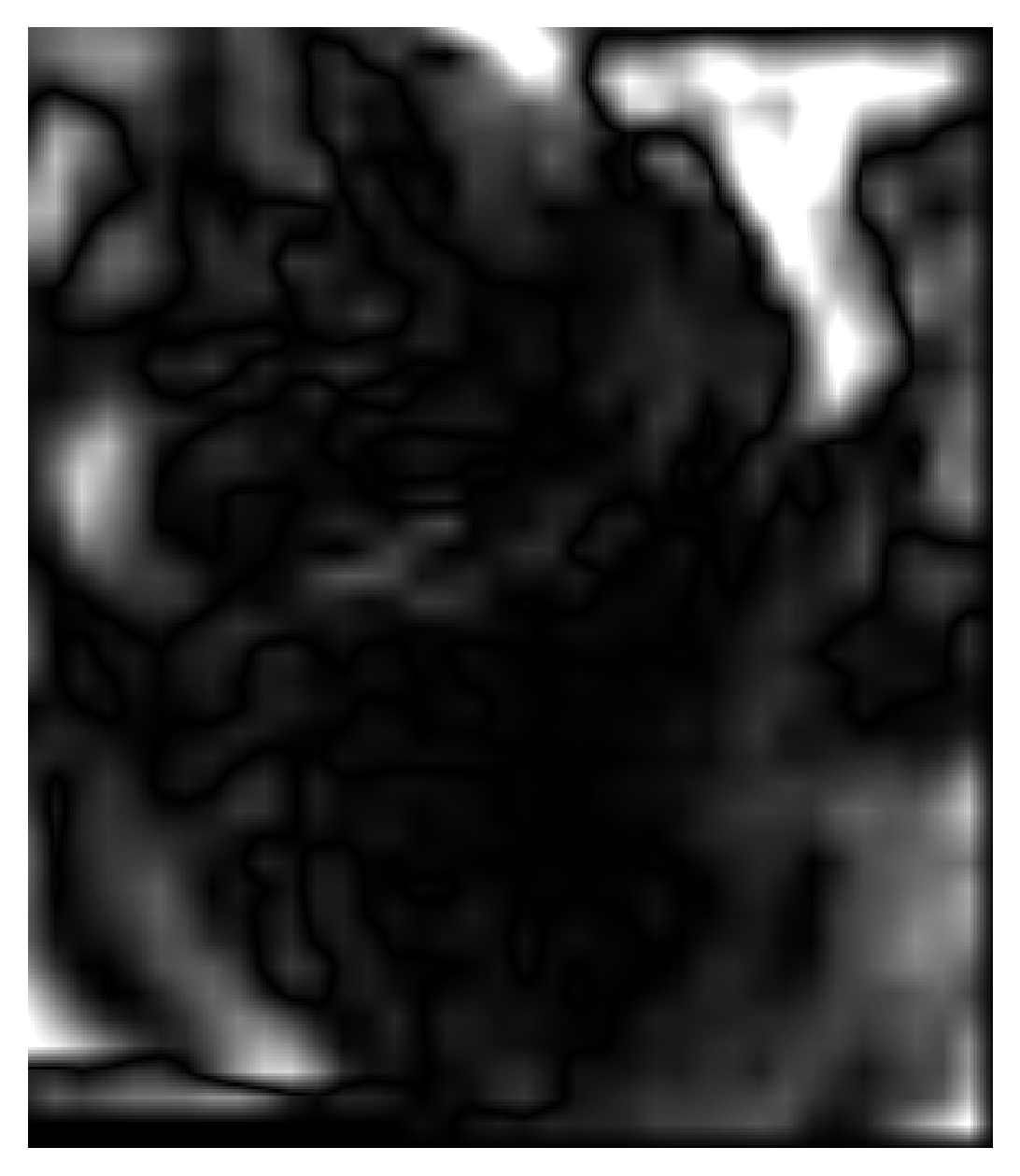}
		\end{minipage}
		\begin{minipage}{0.19\linewidth}
			\includegraphics[width=\textwidth]{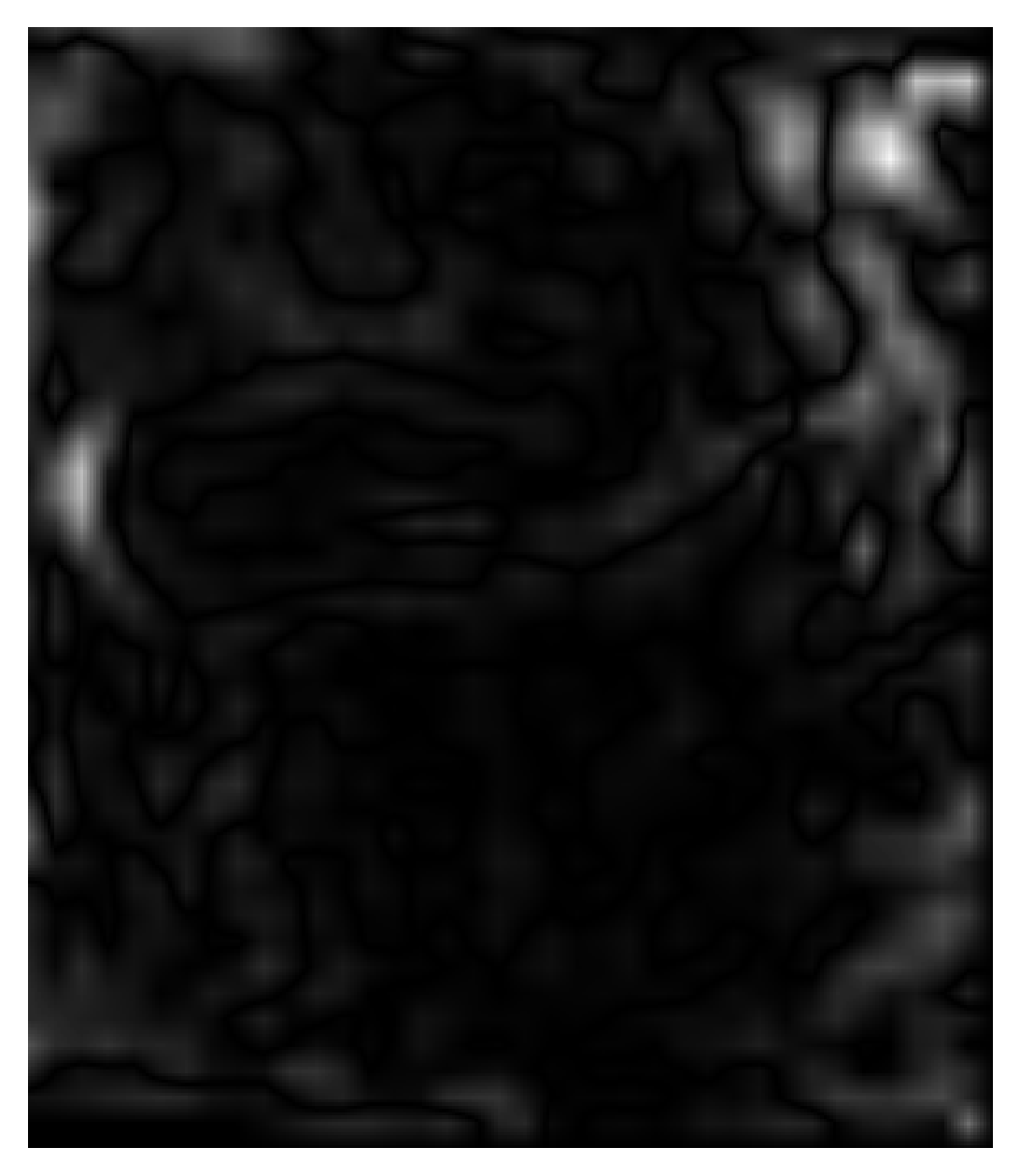}
		\end{minipage}
		\begin{minipage}{0.19\linewidth}
			\includegraphics[width=\textwidth]{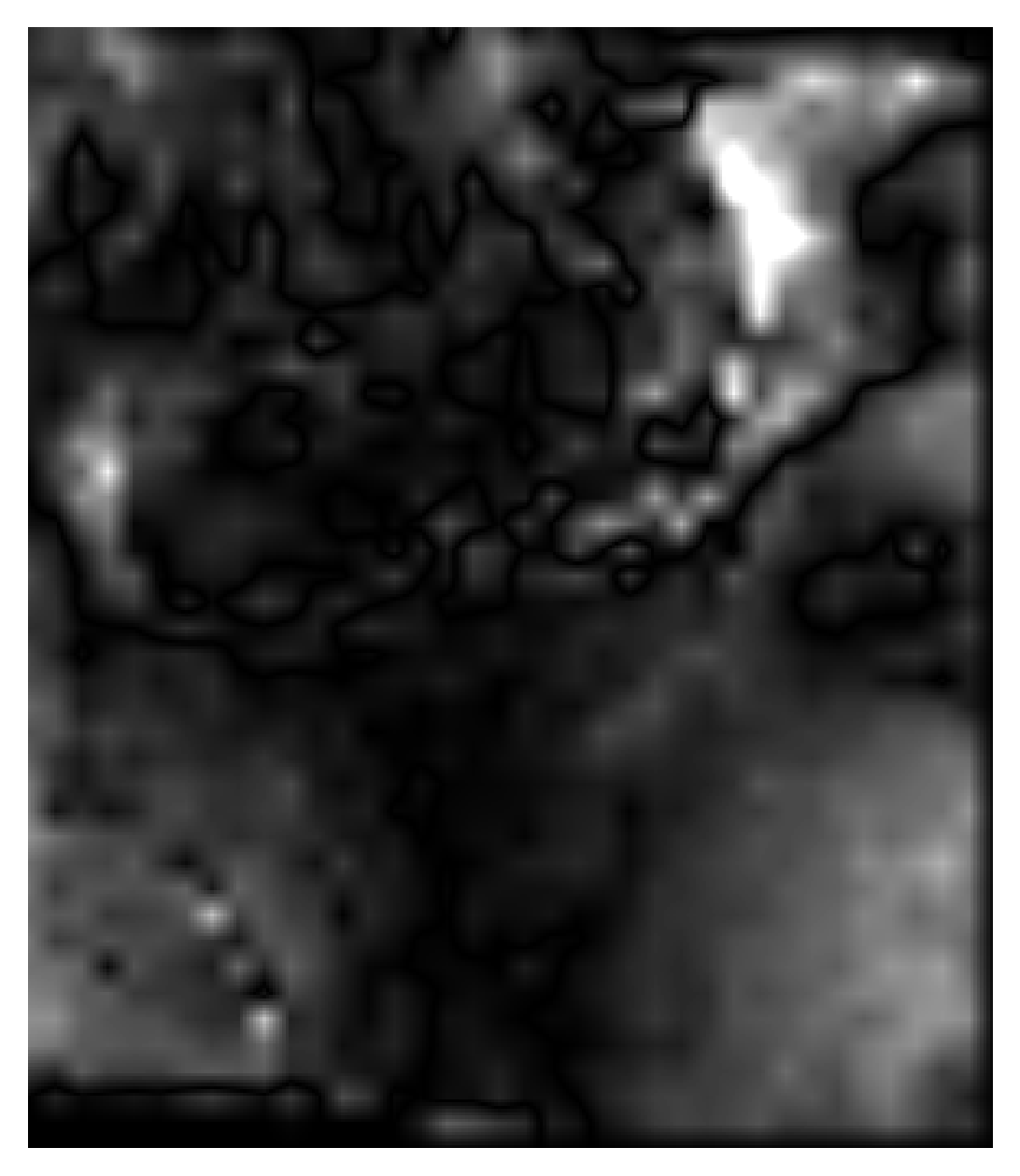}
		\end{minipage}
		\begin{minipage}{0.19\linewidth}
			\includegraphics[width=\textwidth]{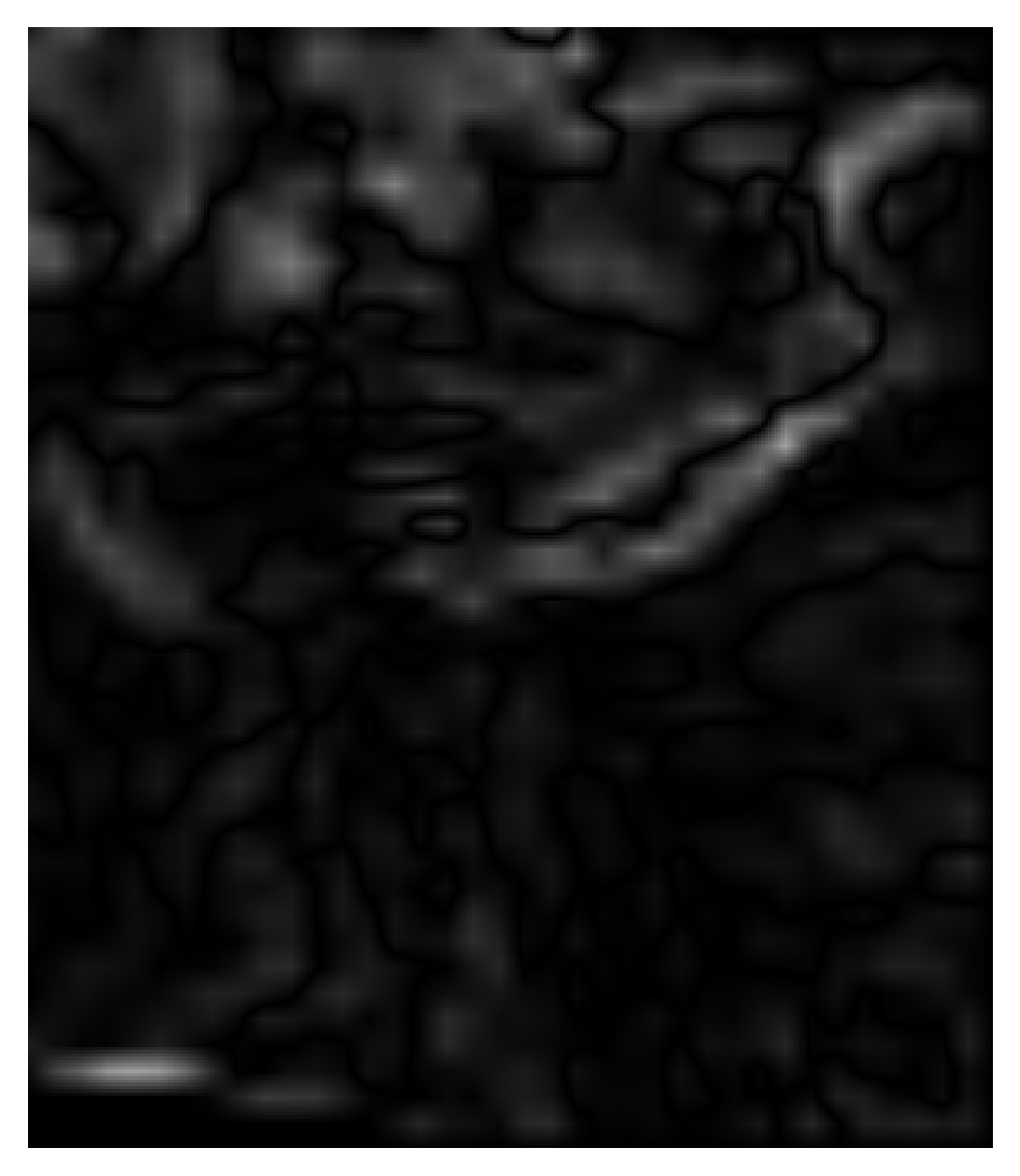}
		\end{minipage}
\end{center}
   \caption{Visualization of registration results from ANTs, NiftyReg, VoxelMorph and ours on the Hippocampus test dataset. The images in the first column are original fixed/moving images with mesh denoting the registration field ground truth. The images from the second column to the last column are reconstructed images with estimated registration field and difference images from ANTs, NiftyReg, VoxelMorph and ours respectively. The first two rows are from axial direction and the last two rows are from sagittal direction. NeurReg obtains the best registration field estimation and smooth difference images with the least error.}
\label{fig:reg_hippo}
\end{figure}

\begin{figure}[t]
\begin{center}
        \begin{minipage}{0.19\linewidth}
			\includegraphics[width=\textwidth]{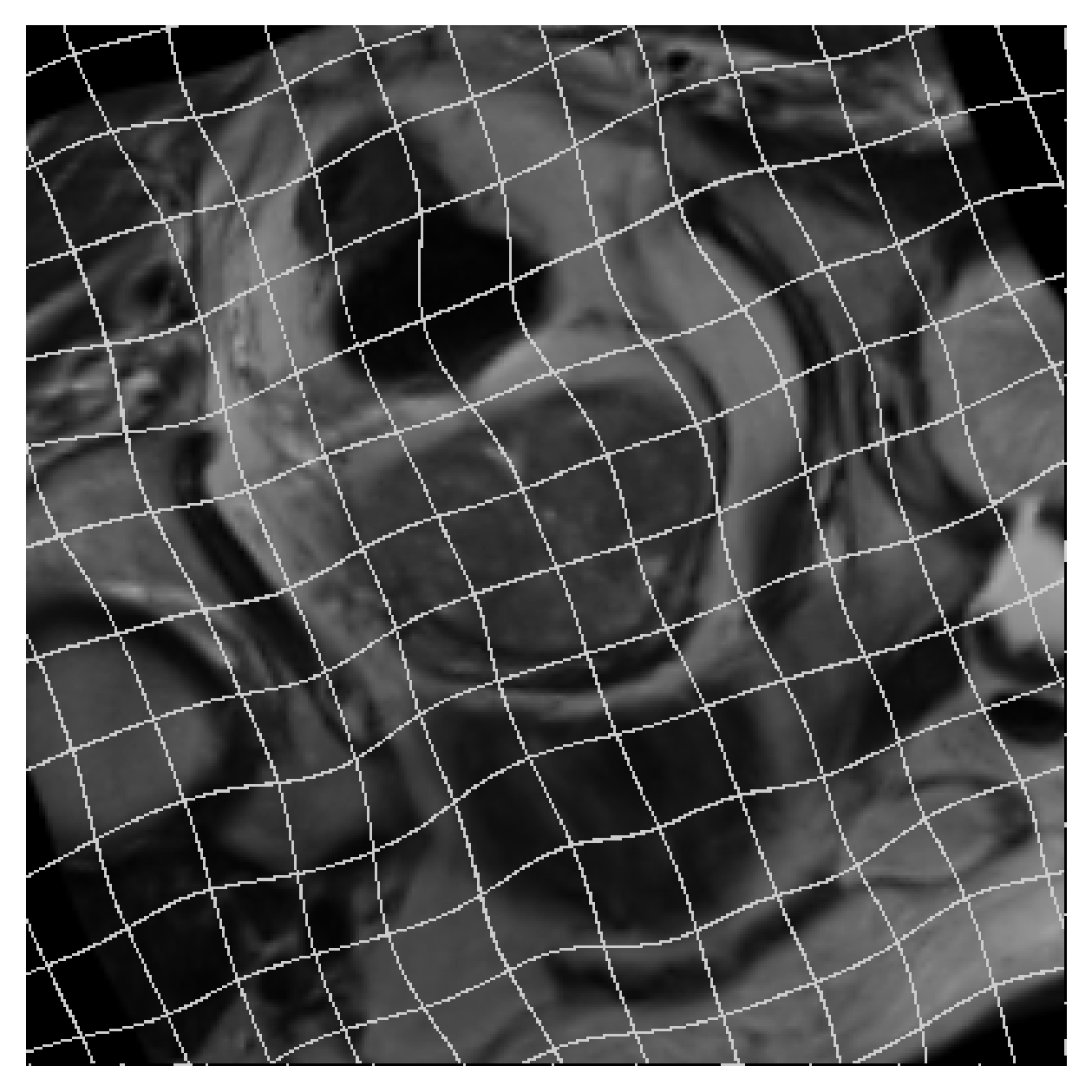}
		\end{minipage}
		\begin{minipage}{0.19\linewidth}
			\includegraphics[width=\textwidth]{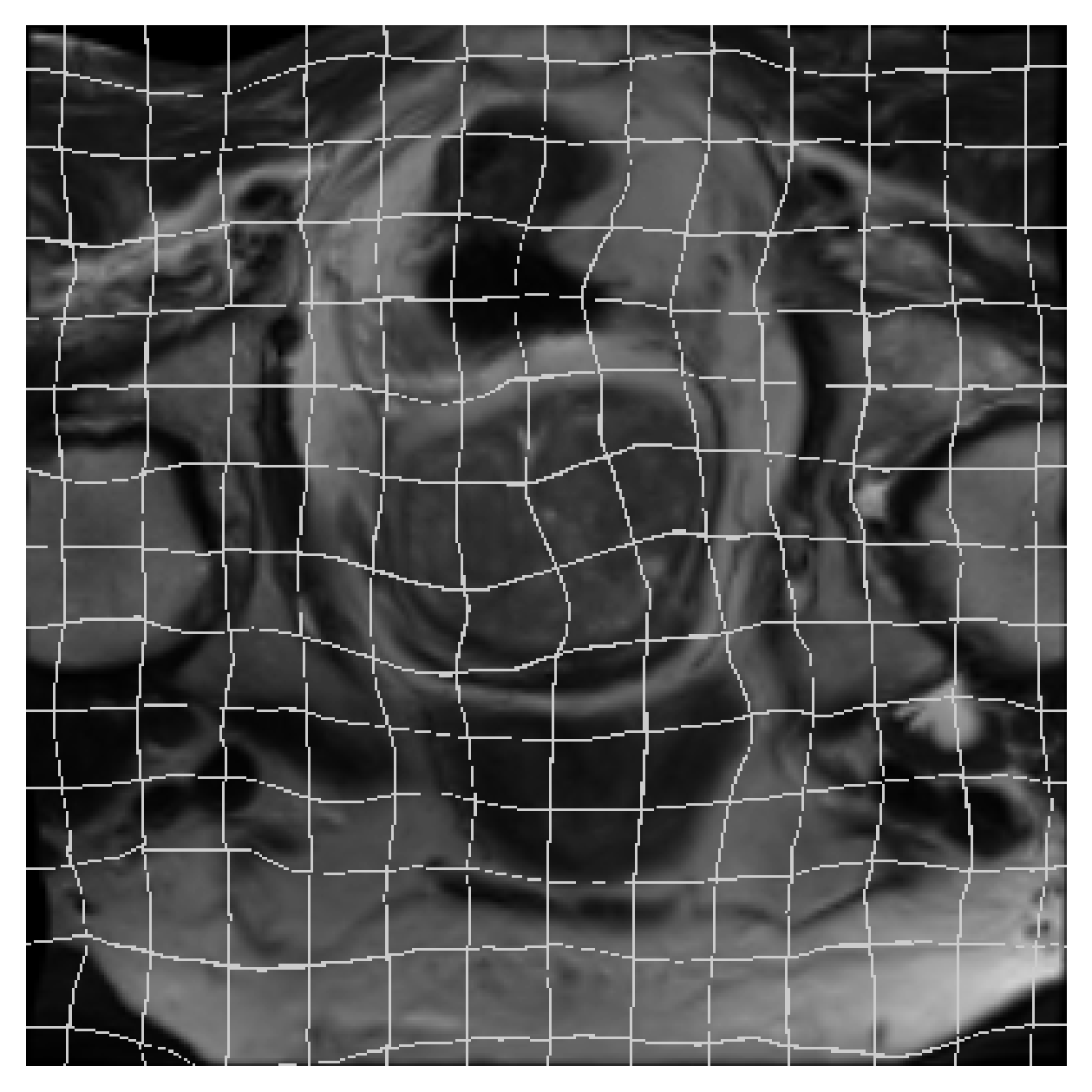}
		\end{minipage}
		\begin{minipage}{0.19\linewidth}
			\includegraphics[width=\textwidth]{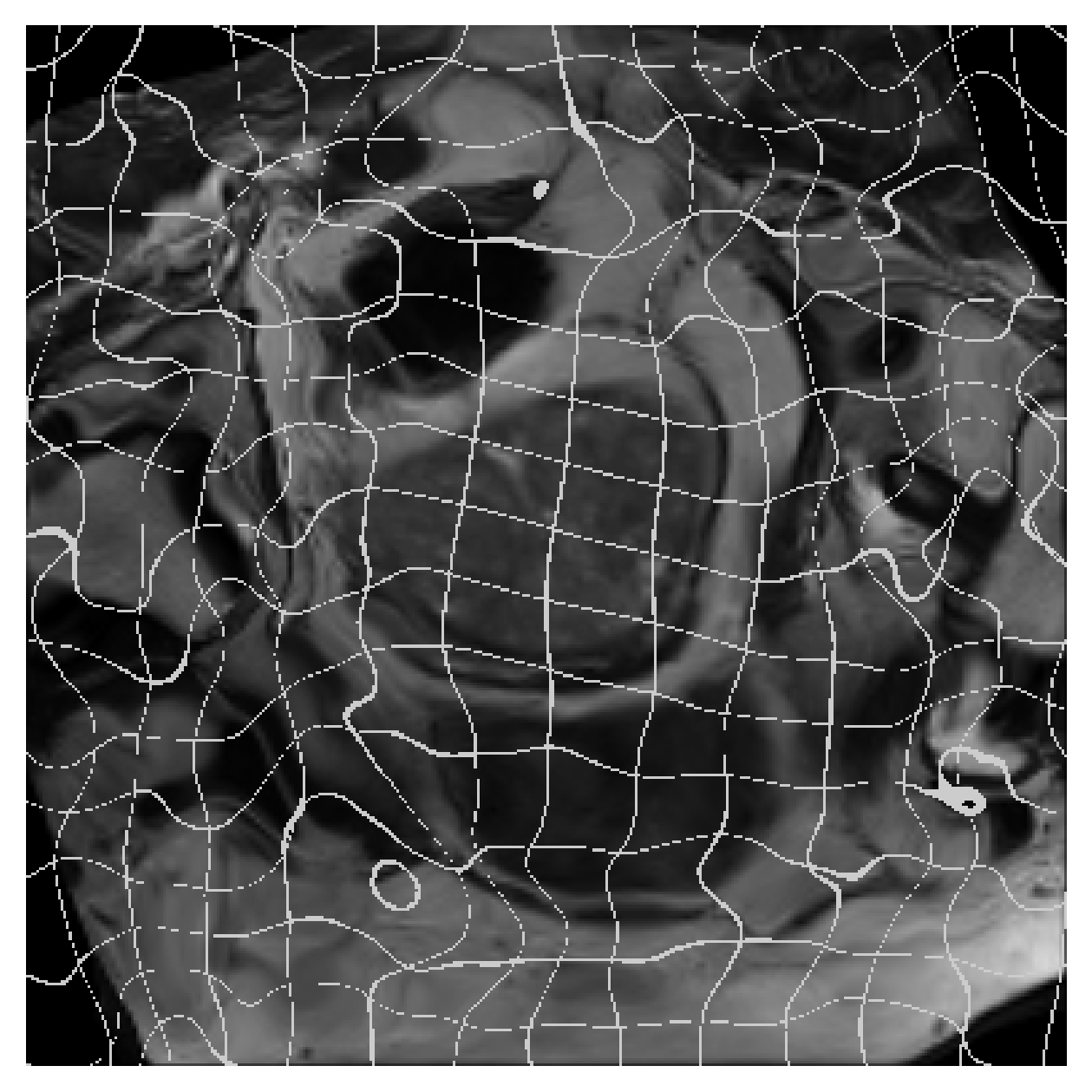}
		\end{minipage}
		\begin{minipage}{0.19\linewidth}
			\includegraphics[width=\textwidth]{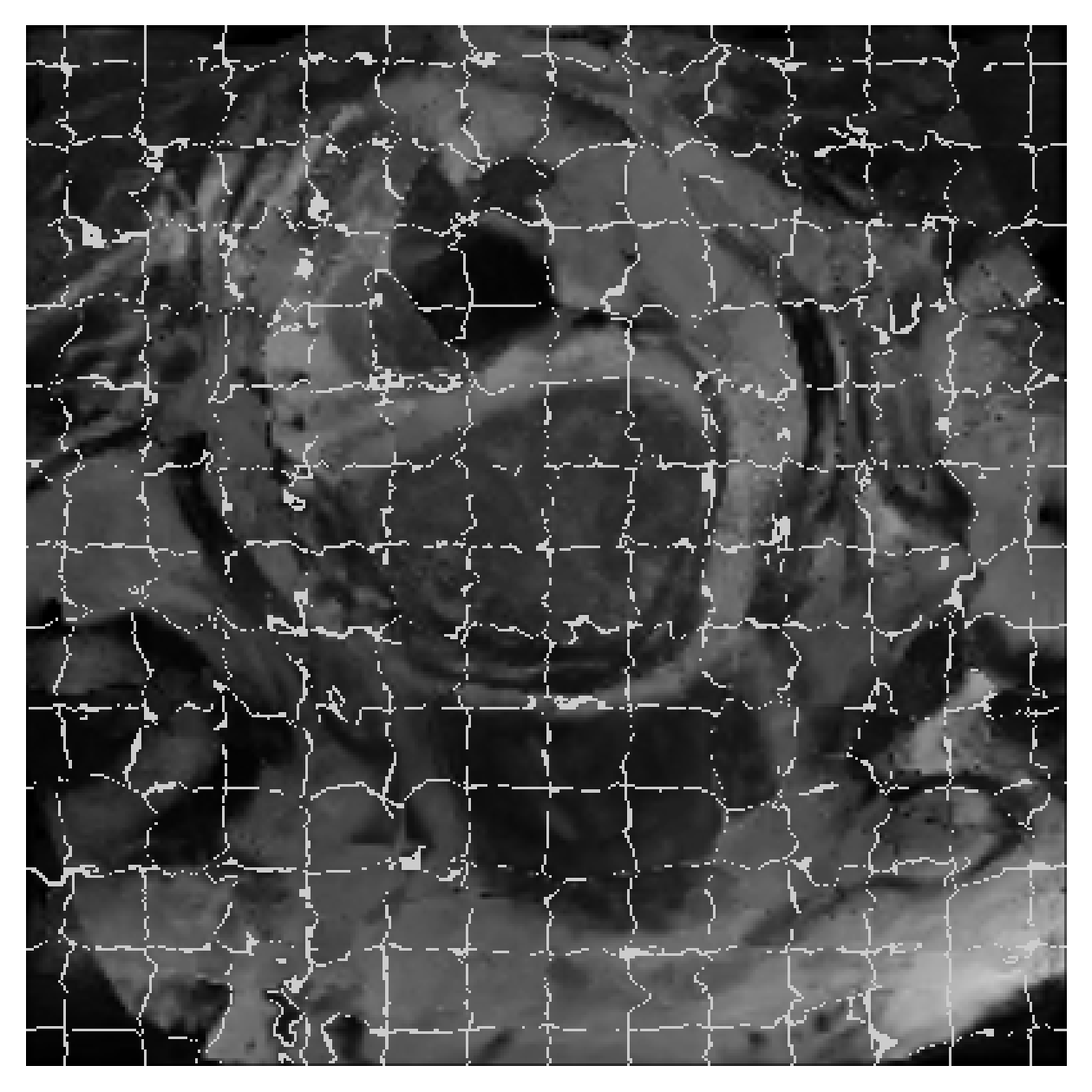}
		\end{minipage}
		\begin{minipage}{0.19\linewidth}
			\includegraphics[width=\textwidth]{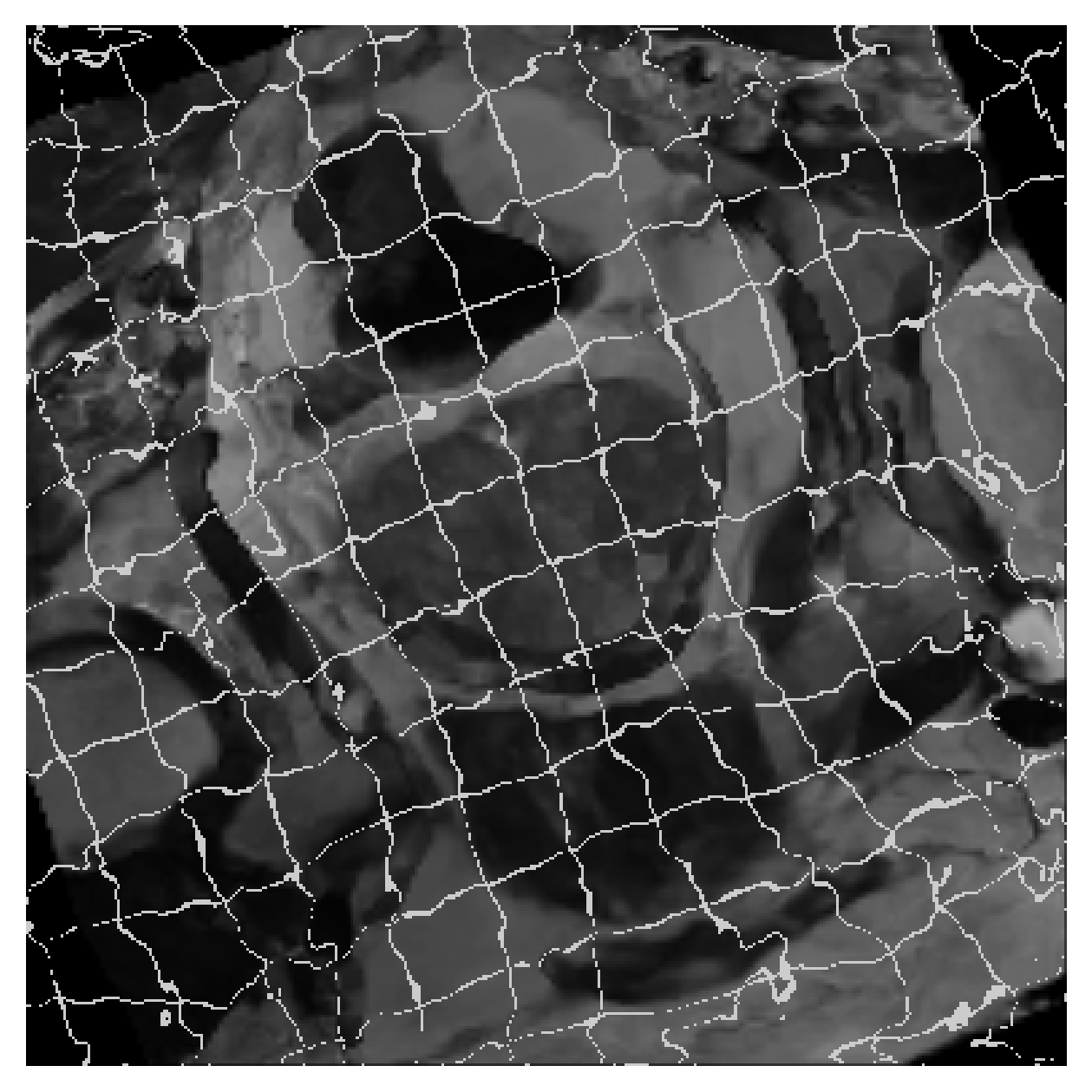}
		\end{minipage}\\
		
		\begin{minipage}{0.19\linewidth}
			\includegraphics[width=\textwidth]{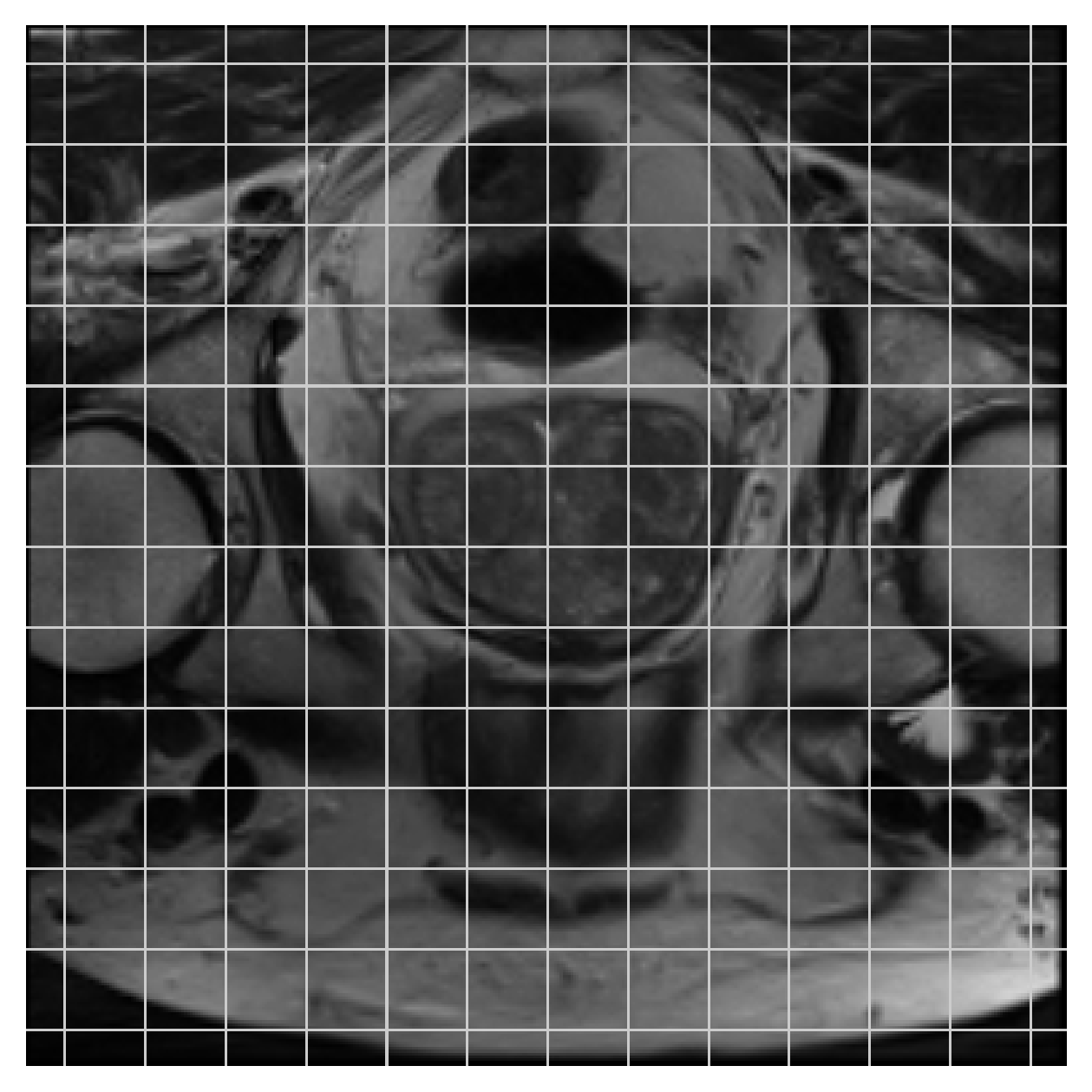}
		\end{minipage}
		\begin{minipage}{0.19\linewidth}
			\includegraphics[width=\textwidth]{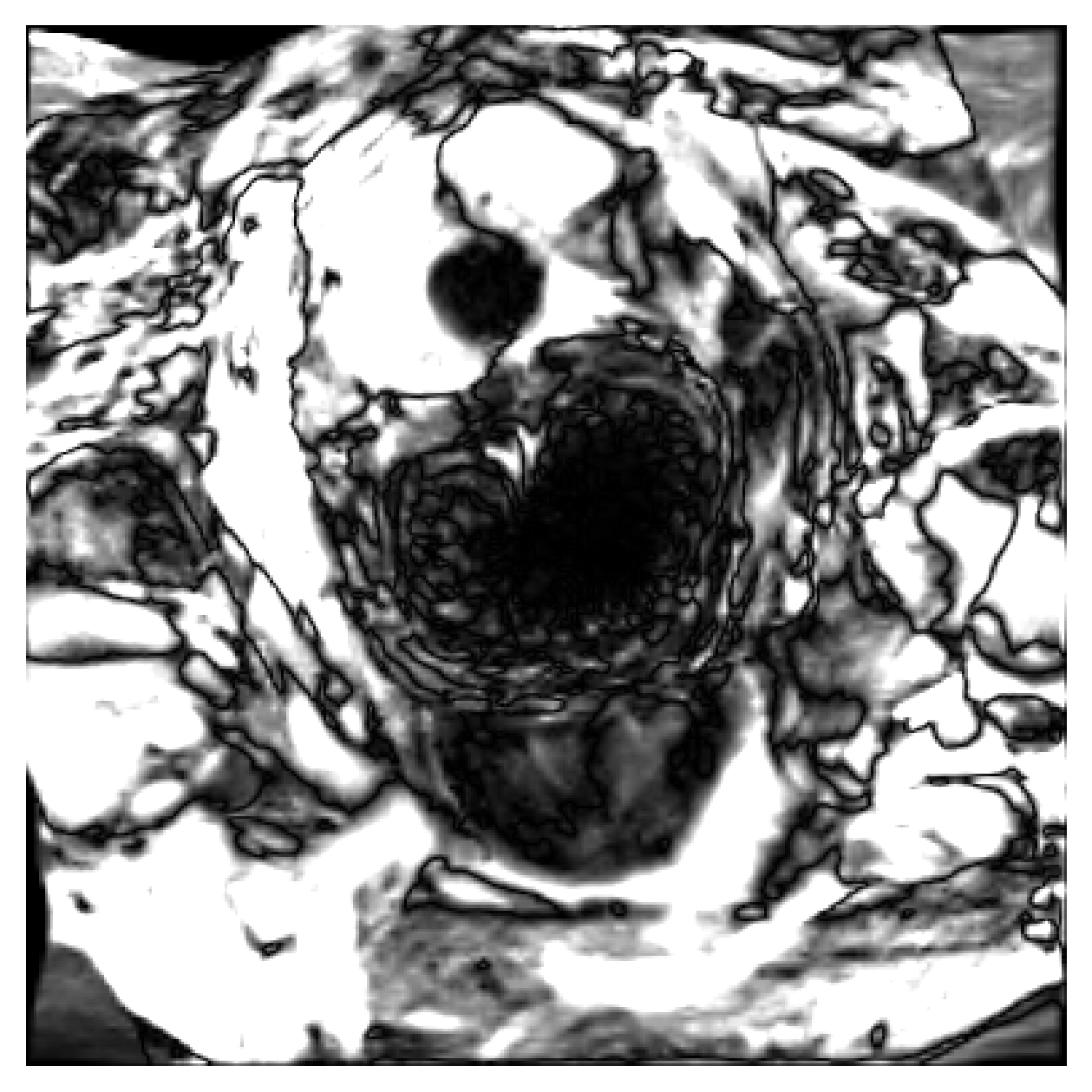}
		\end{minipage}
		\begin{minipage}{0.19\linewidth}
			\includegraphics[width=\textwidth]{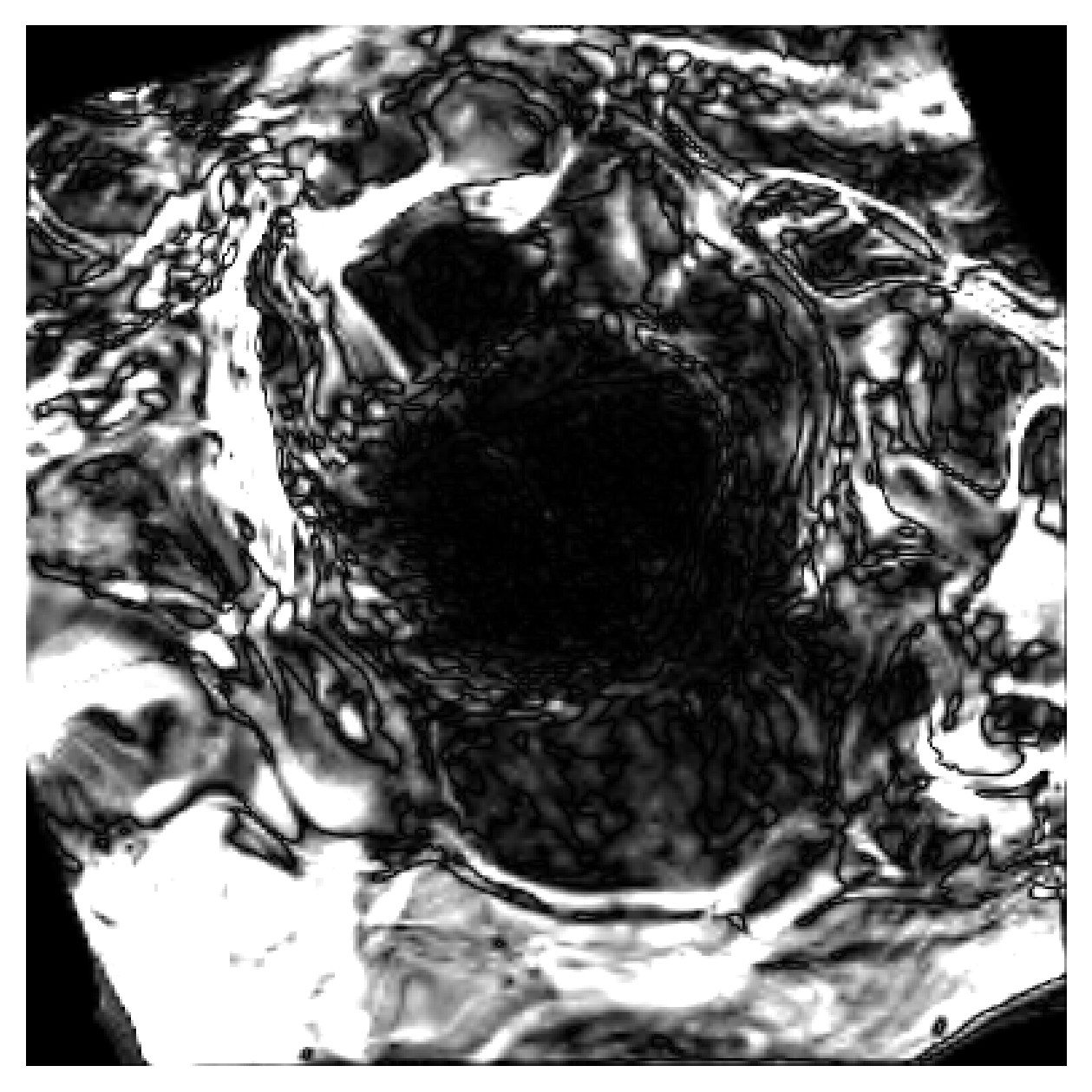}
		\end{minipage}
		\begin{minipage}{0.19\linewidth}
			\includegraphics[width=\textwidth]{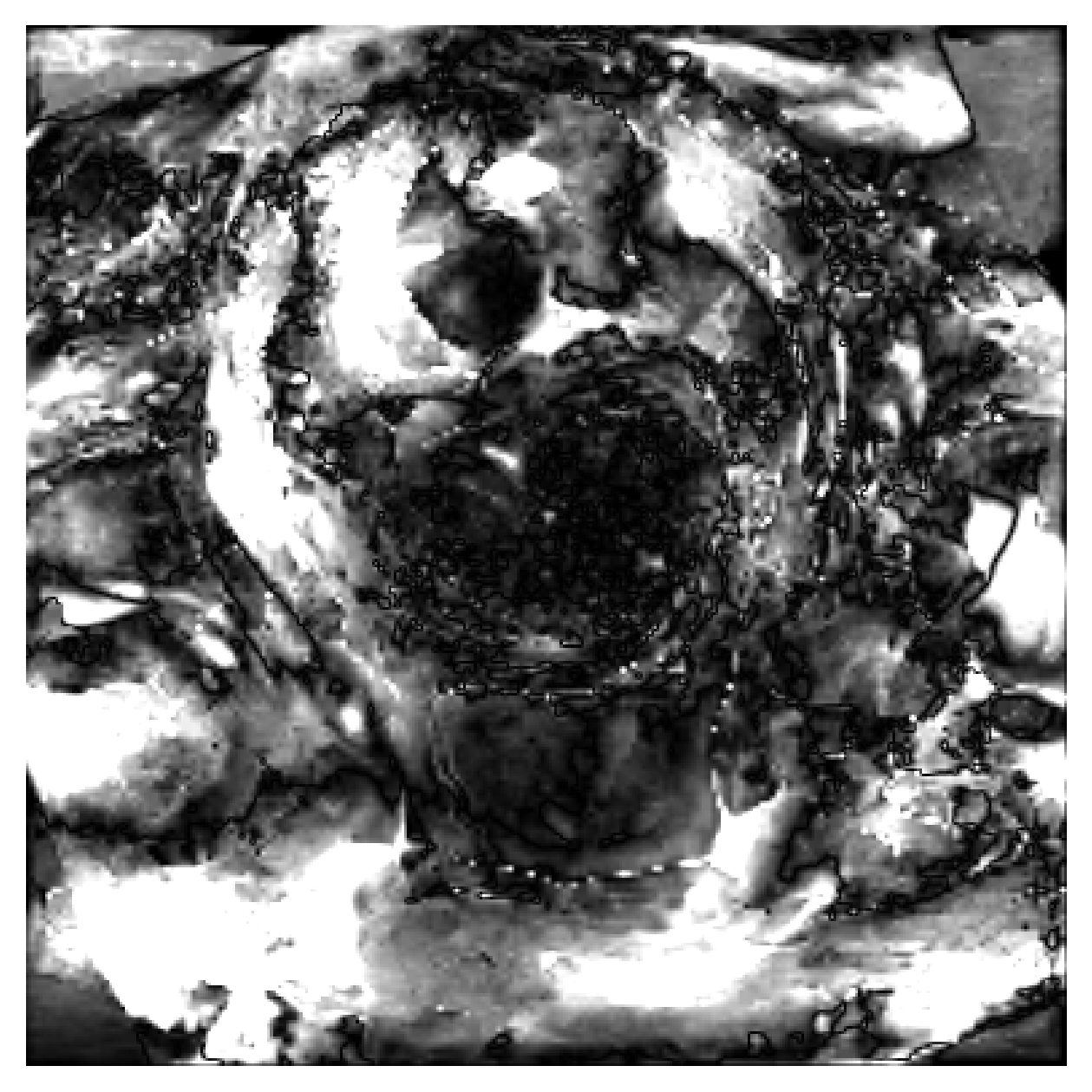}
		\end{minipage}
		\begin{minipage}{0.19\linewidth}
			\includegraphics[width=\textwidth]{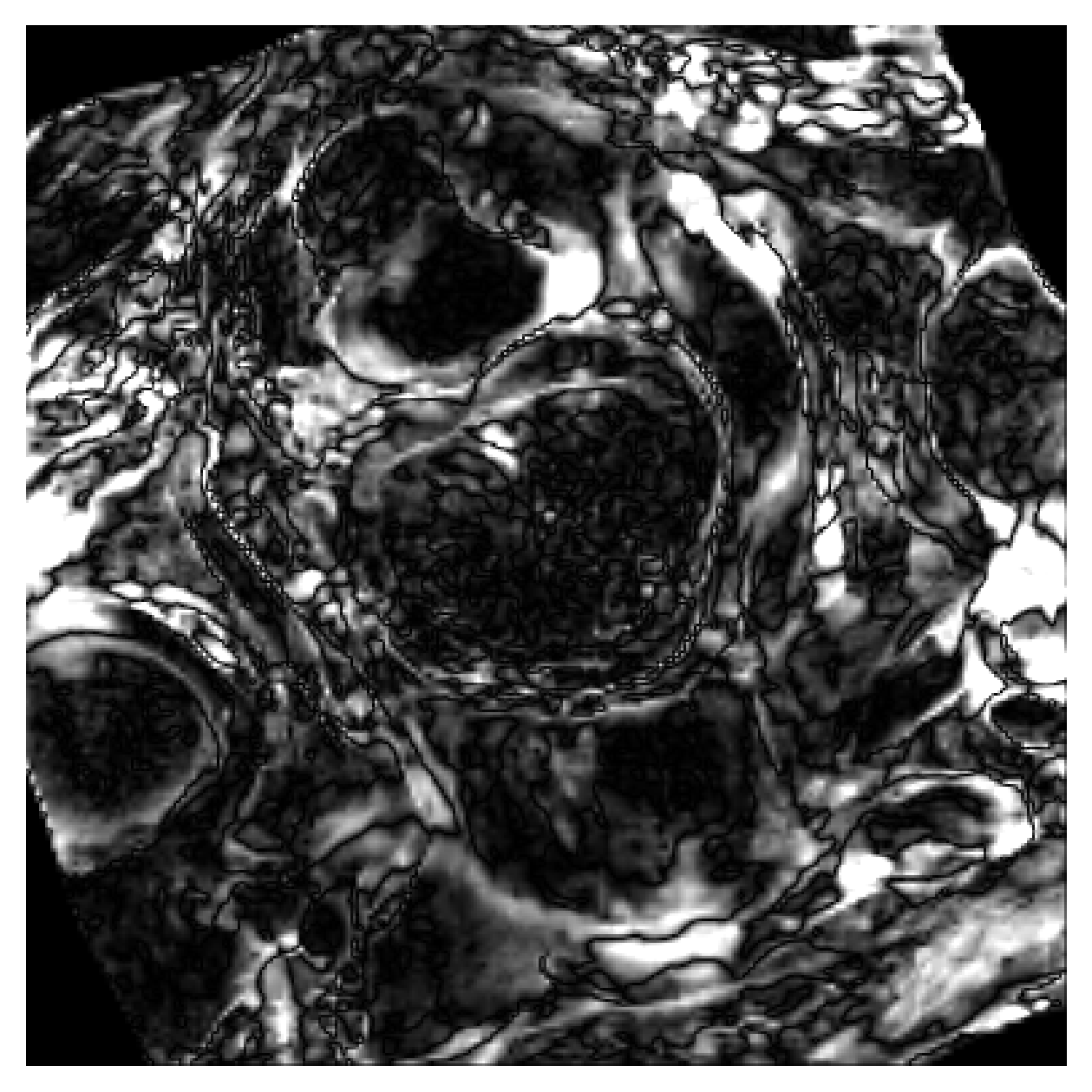}
		\end{minipage}\\
		
		\begin{minipage}{0.19\linewidth}
			\includegraphics[width=\textwidth]{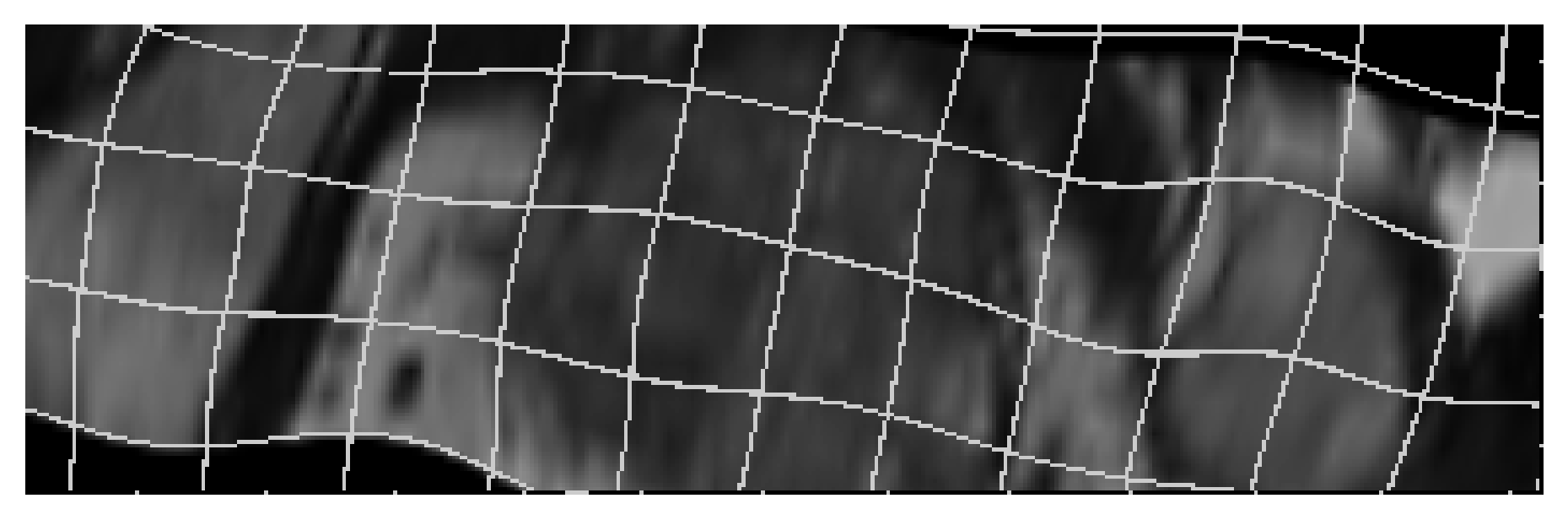}
		\end{minipage}
		\begin{minipage}{0.19\linewidth}
			\includegraphics[width=\textwidth]{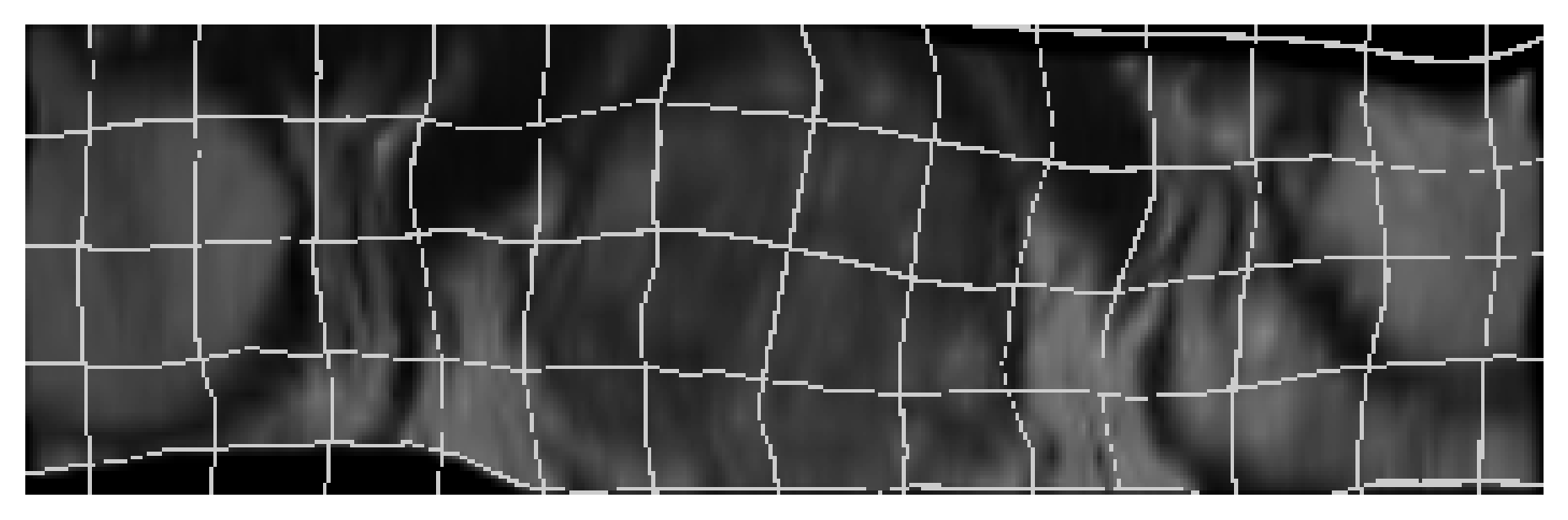}
		\end{minipage}
		\begin{minipage}{0.19\linewidth}
			\includegraphics[width=\textwidth]{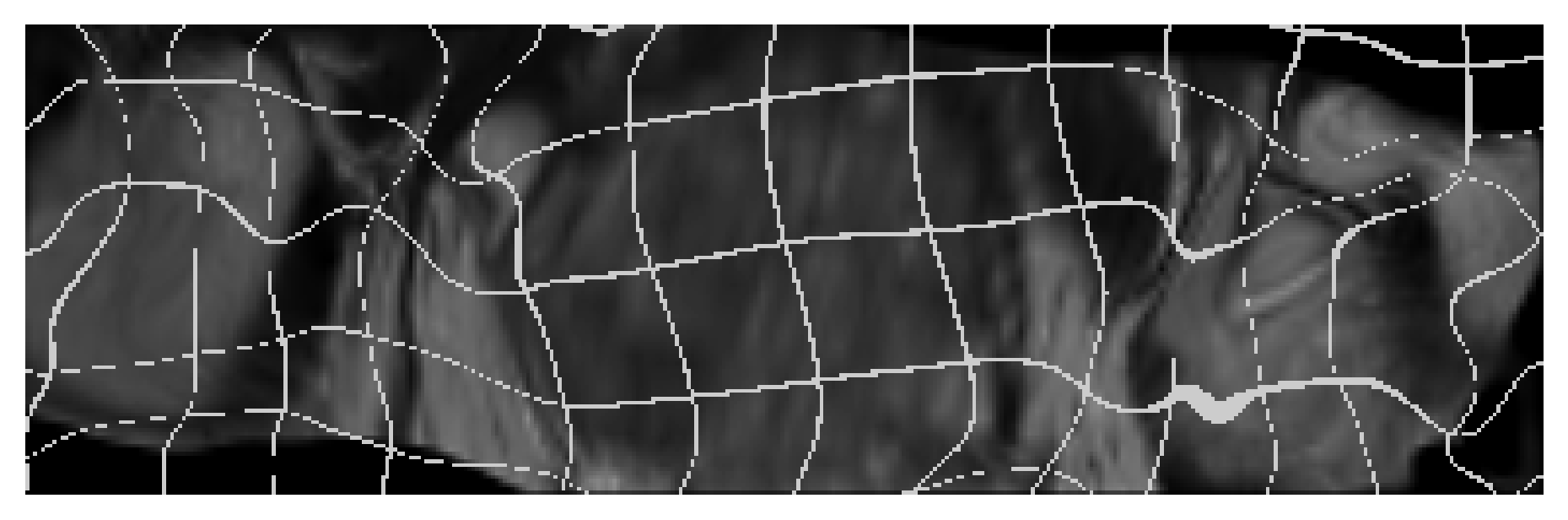}
		\end{minipage}
		\begin{minipage}{0.19\linewidth}
			\includegraphics[width=\textwidth]{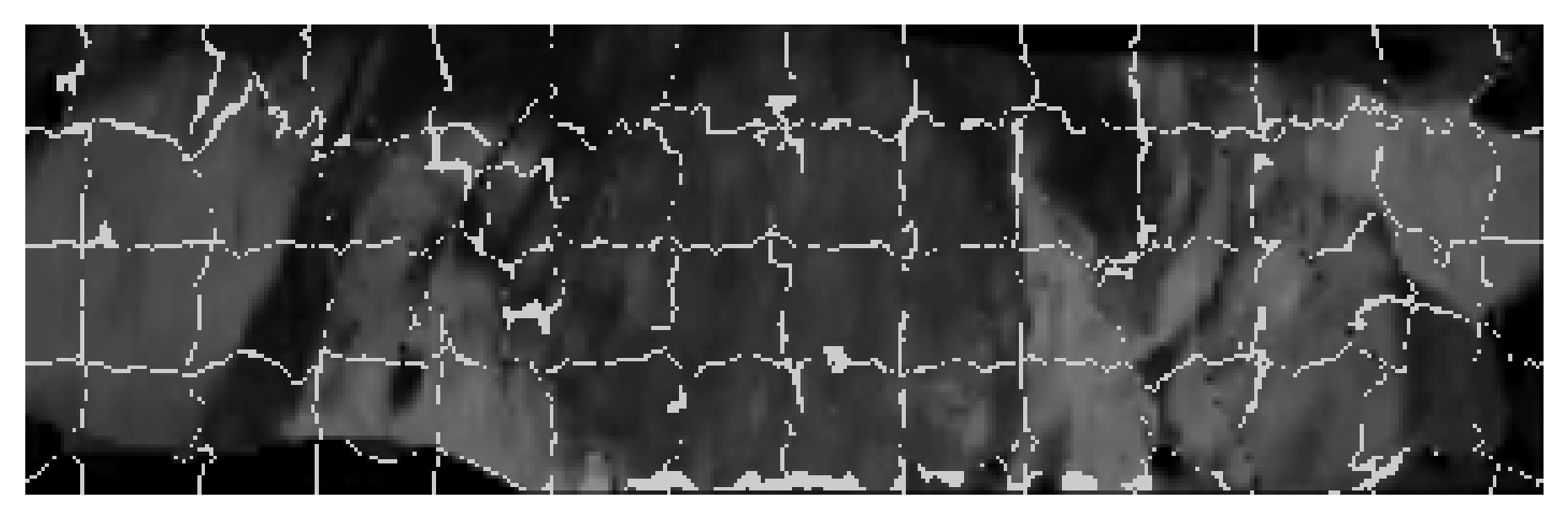}
		\end{minipage}
		\begin{minipage}{0.19\linewidth}
			\includegraphics[width=\textwidth]{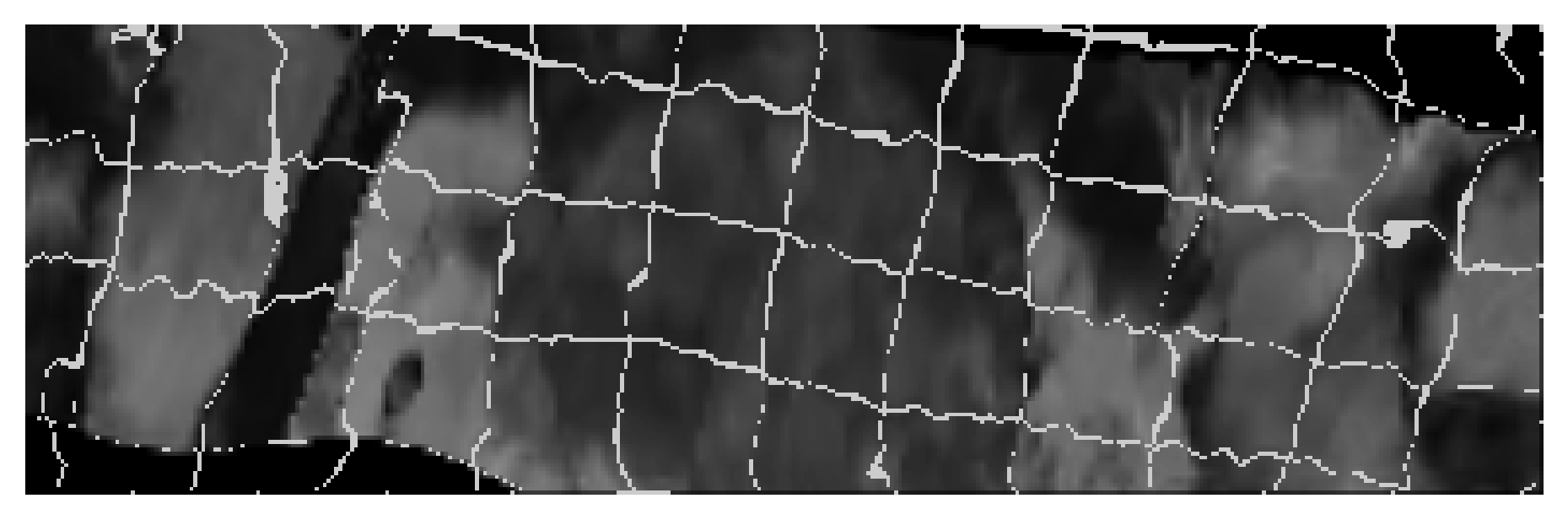}
		\end{minipage}\\
		
		\begin{minipage}{0.19\linewidth}
			\includegraphics[width=\textwidth]{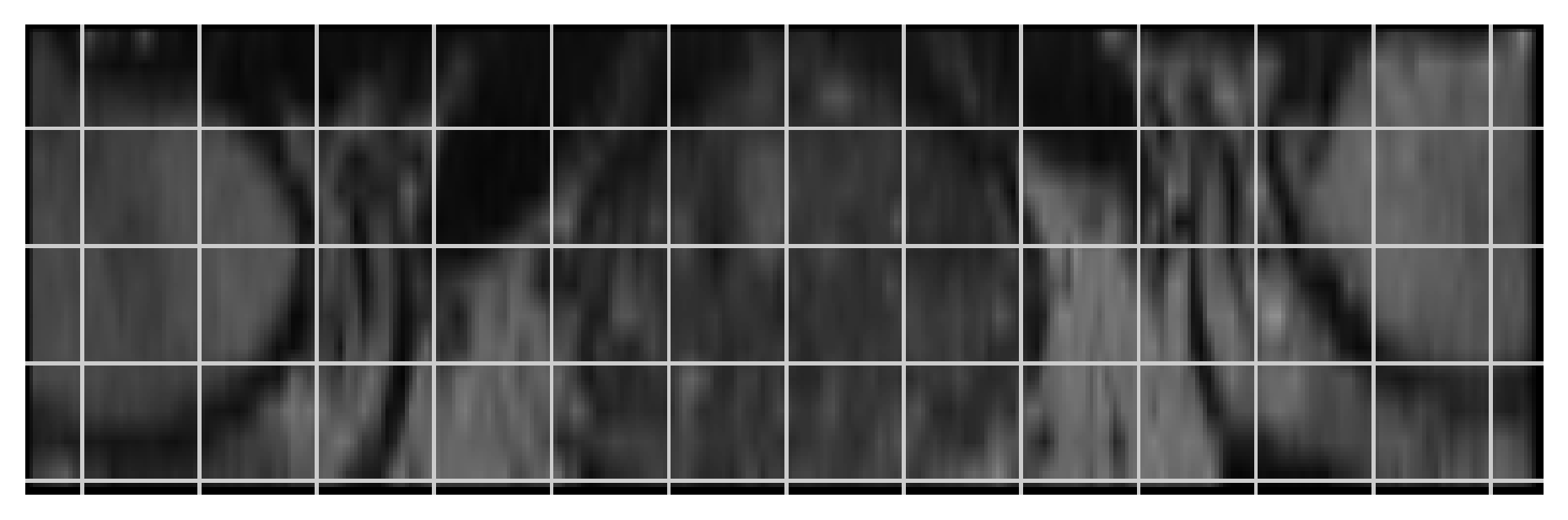}
		\end{minipage}
		\begin{minipage}{0.19\linewidth}
			\includegraphics[width=\textwidth]{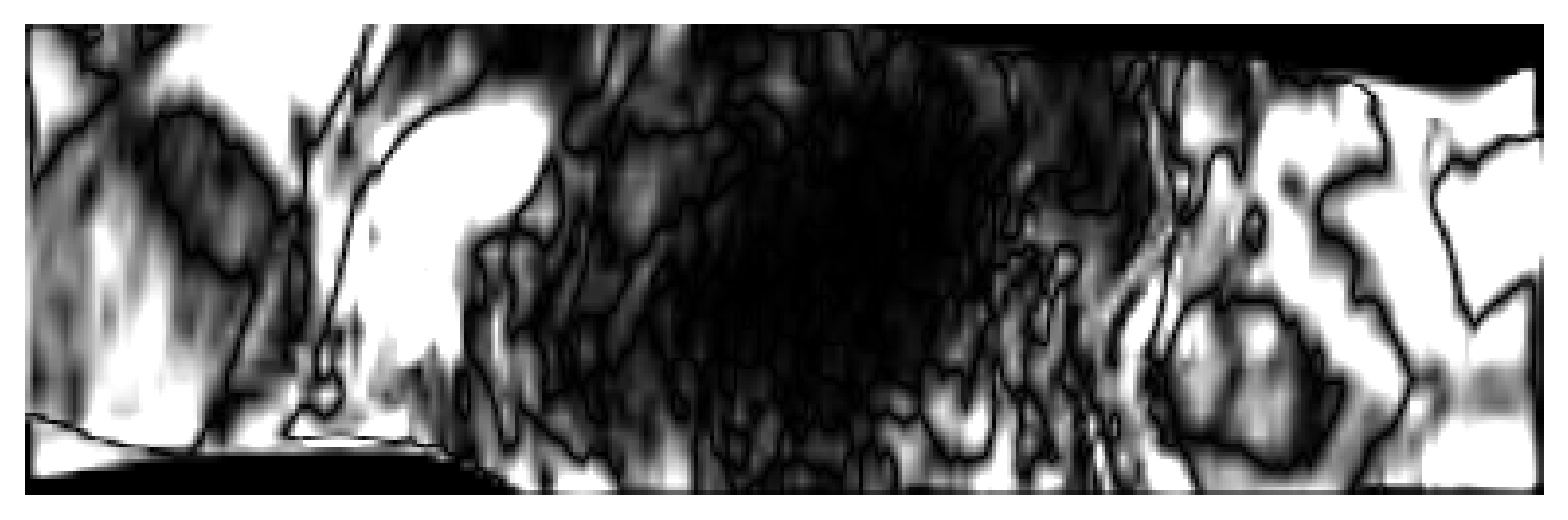}
		\end{minipage}
		\begin{minipage}{0.19\linewidth}
			\includegraphics[width=\textwidth]{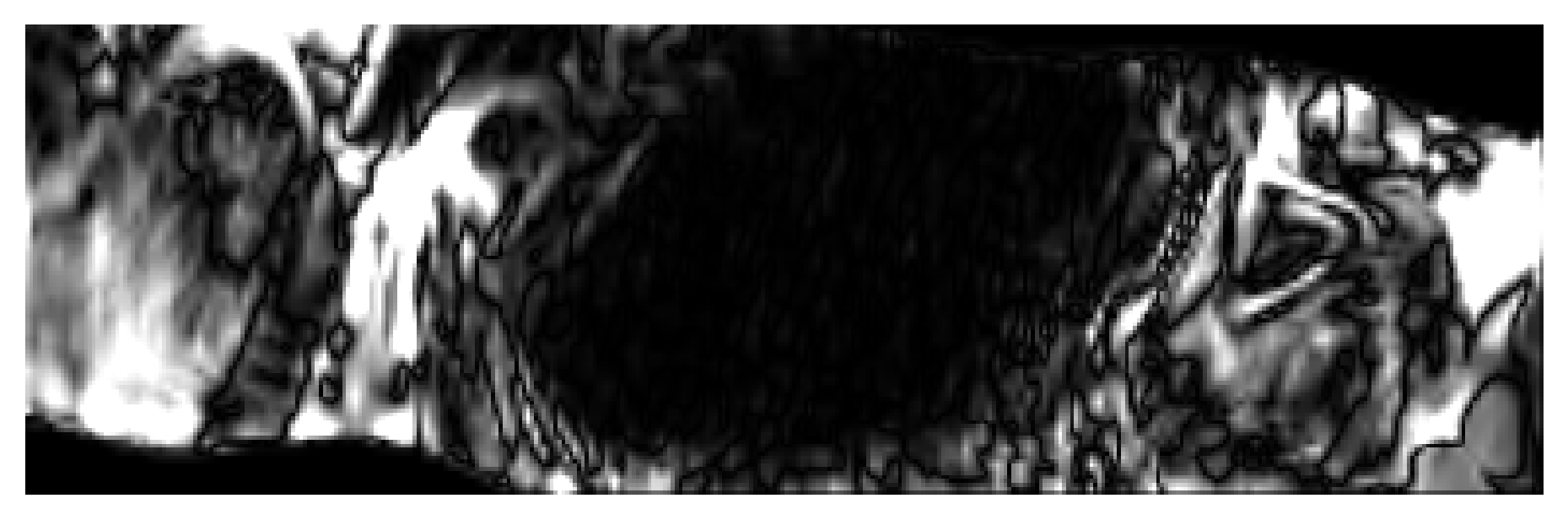}
		\end{minipage}
		\begin{minipage}{0.19\linewidth}
			\includegraphics[width=\textwidth]{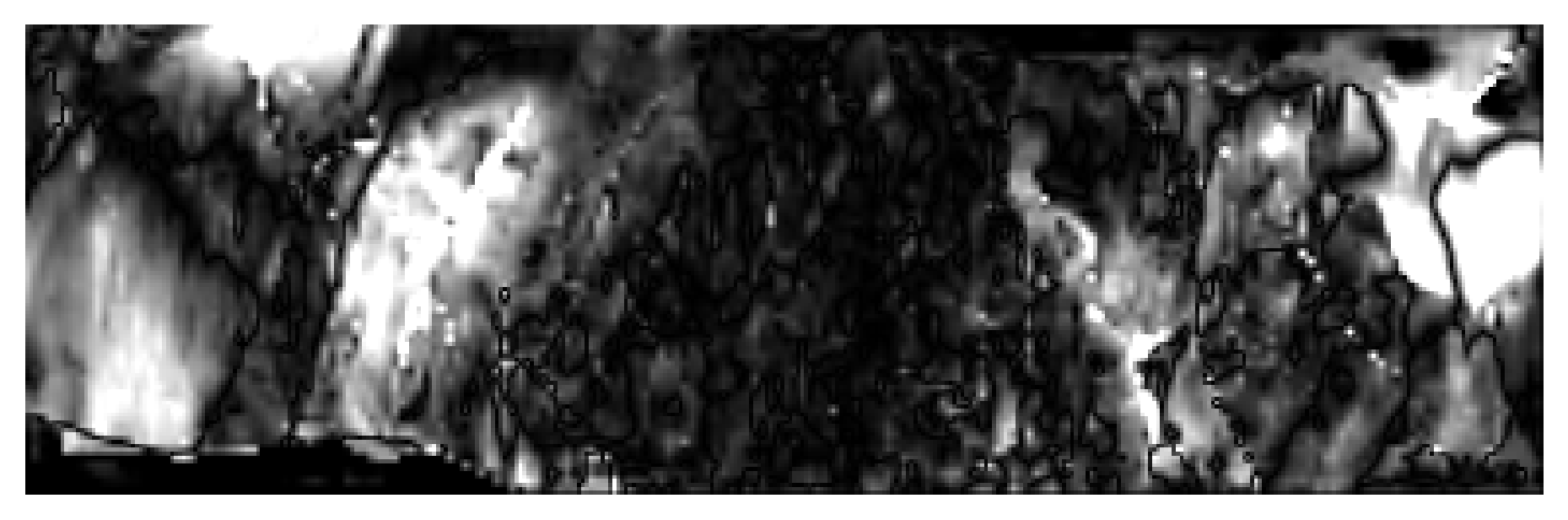}
		\end{minipage}
		\begin{minipage}{0.19\linewidth}
			\includegraphics[width=\textwidth]{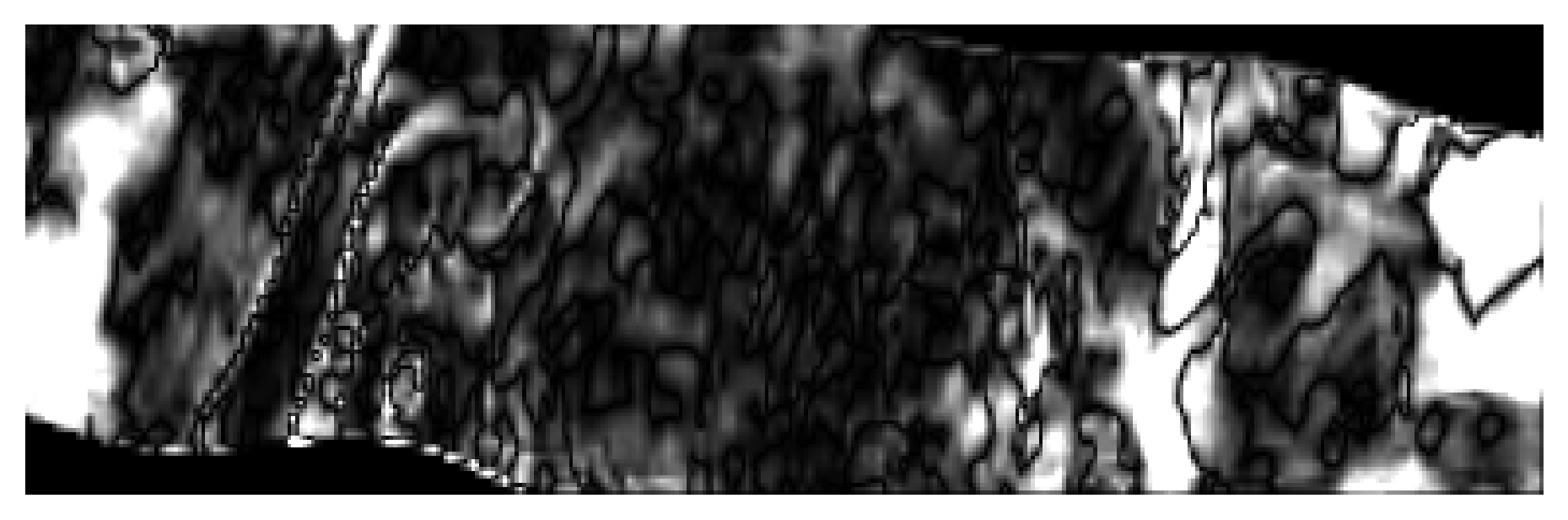}
		\end{minipage}
\end{center}
   \caption{Visualization of registration results from ANTs, NiftyReg, VoxelMorph and ours on the Prostate test datasets. NeurReg obtains the best registration field and difference images.}
\label{fig:reg_pros}
\end{figure}

We further qualitatively compare the four registration methods by visualizing the registration field, reconstructed image and difference image between reconstructed image and fixed image in Fig.~\ref{fig:reg_hippo} and~\ref{fig:reg_pros}. We randomly choose two test cases from the two datasets respectively. For the difference image visualization, we multiply six on the pixel value to increase the intensity for visual purpose.

From the Fig.~\ref{fig:reg_hippo} and~\ref{fig:reg_pros}, NeurReg performs much better than ANTs, NiftyReg and VoxelMorph on the registration field estimation. The difference images of neural registration are smooth without sharp large error points and the overall error is smaller. The ANTs, NiftyReg and VoxelMorph directly optimize the similarity loss which can be considered as the difference image. We introduce the registration field guided supervision and the model converges a better solution for both registration field estimation and appearance reconstruction.

\begin{figure}[t]
\begin{center}
        \begin{minipage}{0.48\linewidth} 
			\includegraphics[width=\textwidth]{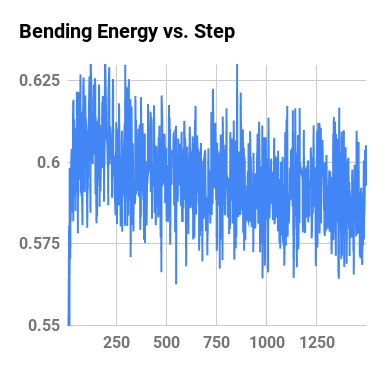}
		\end{minipage}
		\begin{minipage}{0.48\linewidth}
			\includegraphics[width=\textwidth]{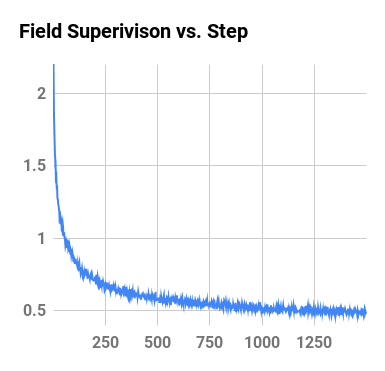}
		\end{minipage}
\end{center}
   \caption{Visualization of bending energy used in VoxelMorph (left) and registration field supervision loss (right) in our registration on the Hippocampus dataset.}
\label{fig:reg_loss}
\end{figure}
We visualize the field training loss of VoxelMorph and NeurReg to further analysis the learning/optimization on the Hippocampus dataset in Fig.~\ref{fig:reg_loss}. We can find the registration field supervision loss is easier to optimize than bending energy used in VoxelMorph and other registrations.
\subsection{Segmentation Performance Comparisons}
We use Dice coefficient (Dice = ${\frac{2TP}{2TP + FN + FP}}$) as the evaluation metric, where TP, FN, and FP are true positives, false negatives, false positives, respectively. For segmentation based on VoxelMorph and NeurReg, we use 10\% training images based on NLCC of reconstructed image as atlases and conduct majority voting in the multi-atlas based segmentation. Because of heavy computational cost of ANTs and NiftyReg, we only use the top one training image based on NLCC in the atlas set from NeurReg as atlas. The multi-task learning from Eq.~\ref{eq:loss_reg_mtl} is denoted by VoxelMorph (MTL) and Ours (MTL). The residual segmentation block in Fig.~\ref{fig:boost_segm} is denoted by VoxelMorph (Feat.) and Ours (Feat.). We also compare with an advanced segmentation model which uses 3D U-Net based on 18 residual blocks in the encoder and ImageNet pretrained Res-18 weight~\cite{liu20183d}. The comparison results are listed in table~\ref{tab:seg_hippo} and~\ref{tab:seg_pros}.
{\small \begin{table}
\begin{center}
\begin{tabular}{|l|c|}
\hline
Method & Dice (\%)  \\
\hline\hline
ANTs &80.86$\pm$5.13/78.34$\pm$5.24  \\
NiftyReg &80.53$\pm$4.86/77.92$\pm$5.47 \\
VoxelMorph &\begin{tabular}{@{}c@{}}78.92$\pm$6.79/75.85$\pm$8.13\end{tabular} \\
VoxelMorph (MTL) &86.69$\pm$5.04/85.72$\pm$3.97 \\
VoxelMorph (Feat.) &86.68$\pm$4.24/85.68$\pm$4.46  \\
3D U-Net &88.66$\pm$3.09/86.76$\pm$3.26  \\
\hline
Ours &\begin{tabular}{@{}c@{}}78.99$\pm$6.24/78.72$\pm$7.47\end{tabular}   \\
Ours (MTL) &88.95$\pm$3.66/87.10$\pm$3.72   \\
Ours (Feat.) &${\bm{89.18\pm 3.50/87.39\pm 3.42}}$   \\
\hline
\end{tabular}
\end{center}
\caption{Segmentation comparisons on the Hippocampus dataset. Ours is the best and comparable with an advanced 3D UNet.}
\label{tab:seg_hippo}
\end{table}}

{\small \begin{table}
\begin{center}
\begin{tabular}{|l|c|}
\hline
Method & Dice (\%) \\
\hline\hline
ANTs &28.52$\pm$18.94/57.58$\pm$24.09 \\
NiftyReg &27.88$\pm$17.25/56.67$\pm$23.71  \\
VoxelMorph &22.03$\pm$12.33/53.40$\pm$21.50  \\ 
VoxelMorph (MTL) &25.50$\pm$14.63/63.62$\pm$18.73 \\
VoxelMorph (Feat.) &38.97$\pm$16.61/76.72$\pm$5.62  \\
\hline 
Ours &21.70$\pm$11.95/55.80$\pm$17.53 \\  
Ours (MTL) &31.57$\pm$22.49/73.84$\pm$12.68   \\ 
Ours (Feat.) &${\bm{44.30\pm 17.60/82.38\pm 3.46}}$  \\
\hline
\end{tabular}
\end{center}
\caption{Segmentation comparisons on the Prostate dataset.}
\label{tab:seg_pros}
\end{table}}

Table~\ref{tab:seg_hippo} demonstrates better accuracy of our registration-based segmentations compared to VoxelMorph on all the three frameworks. The NeurReg with residual segmentation block obtains comparable performance as the advanced 3D U-Net with more parameters and pretrained model~\cite{liu20183d}. The atlas-based segmentation with fast inference speed is comparable with NiftyReg and ANTs and it confirms the robustness of proposed registration. From table~\ref{tab:seg_pros}, NeurReg with residual segmentation block achieves the best performance which might be because of the dual registration and residual segmentation block.

\begin{figure}[t]
\begin{center}
\begin{minipage}{0.19\linewidth}
	\includegraphics[width=\textwidth]{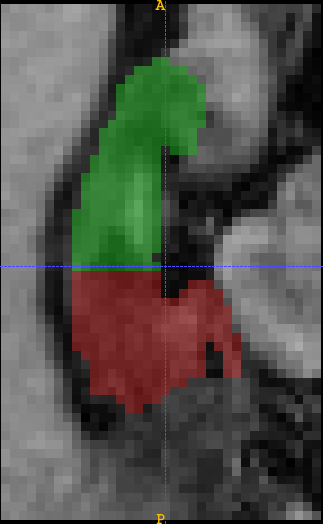}
\end{minipage}
\begin{minipage}{0.19\linewidth}
	\includegraphics[width=\textwidth]{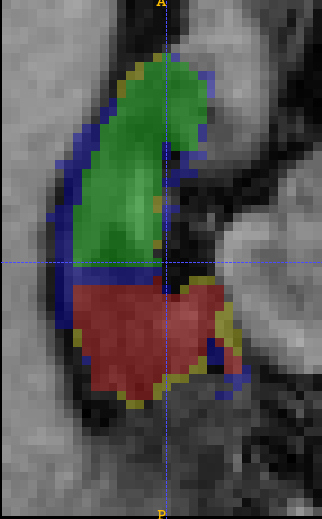}
\end{minipage}
\begin{minipage}{0.19\linewidth}
	\includegraphics[width=\textwidth]{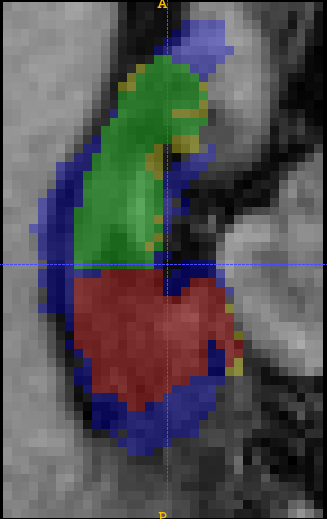}
\end{minipage}
\begin{minipage}{0.19\linewidth}
	\includegraphics[width=\textwidth]{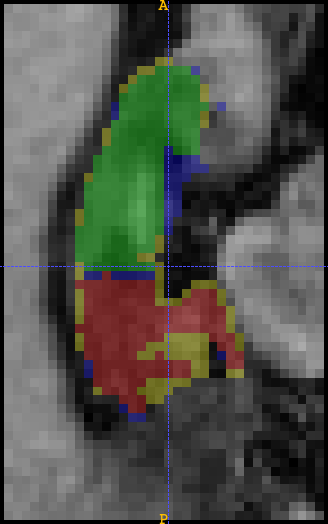}
\end{minipage}
\begin{minipage}{0.19\linewidth}
	\includegraphics[width=\textwidth]{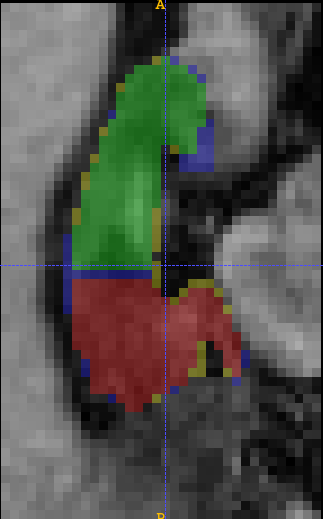}
\end{minipage}\\

\begin{minipage}{0.19\linewidth}
	\includegraphics[width=\textwidth]{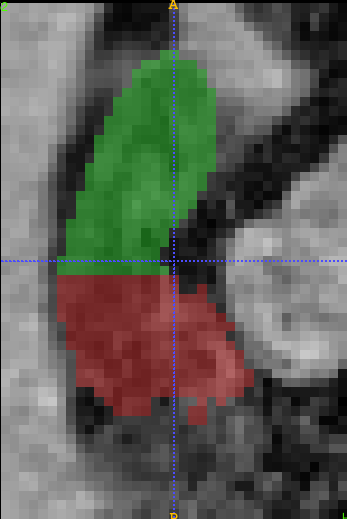}
\end{minipage}
\begin{minipage}{0.19\linewidth}
	\includegraphics[width=\textwidth]{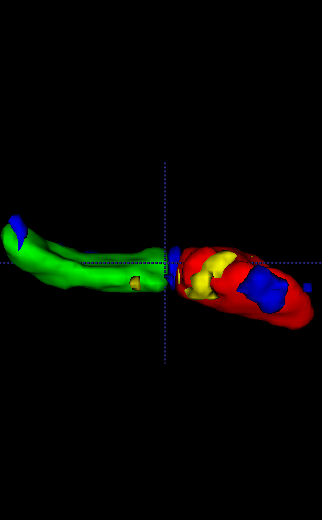}
\end{minipage}
\begin{minipage}{0.19\linewidth}
	\includegraphics[width=\textwidth]{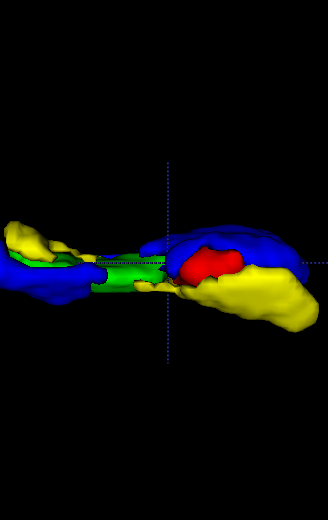}
\end{minipage}
\begin{minipage}{0.19\linewidth}
	\includegraphics[width=\textwidth]{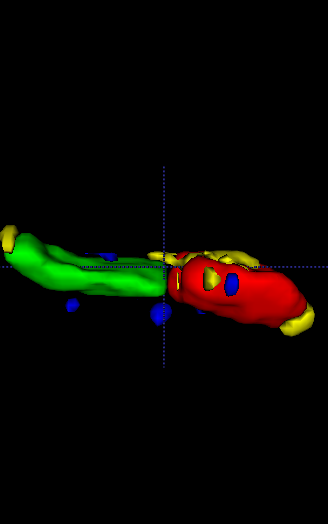}
\end{minipage}
\begin{minipage}{0.19\linewidth}
	\includegraphics[width=\textwidth]{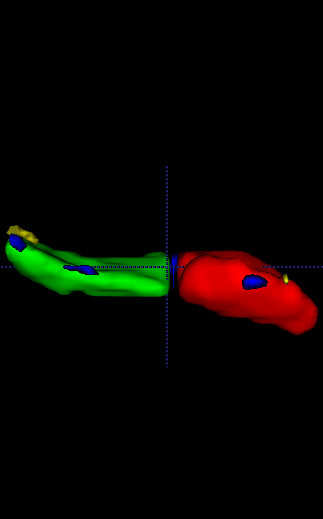}
\end{minipage}\\

\begin{minipage}{0.19\linewidth}
	\includegraphics[width=\textwidth]{./fig/hippo_252_0_15_seg_mov.png}
\end{minipage}
\begin{minipage}{0.19\linewidth}
	\includegraphics[width=\textwidth]{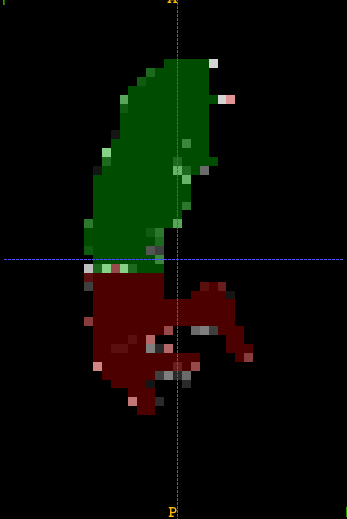}
\end{minipage}
\begin{minipage}{0.19\linewidth}
	\includegraphics[width=\textwidth]{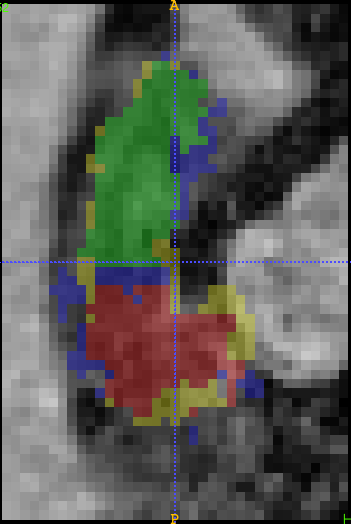}
\end{minipage}
\begin{minipage}{0.19\linewidth}
	\includegraphics[width=\textwidth]{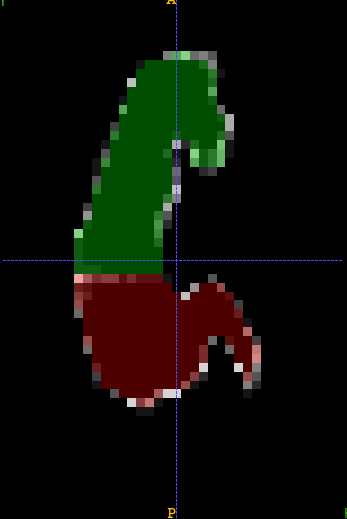}
\end{minipage}
\begin{minipage}{0.19\linewidth}
	\includegraphics[width=\textwidth]{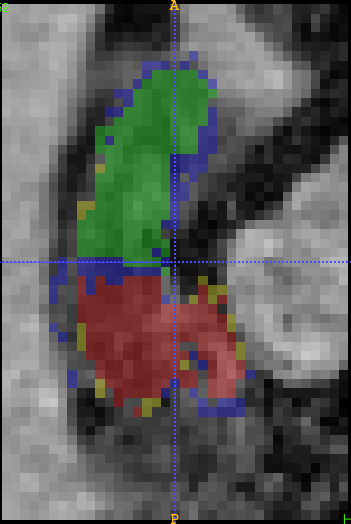}
\end{minipage}
\end{center}
   \caption{Visualization of segmentation results from ANTs, NiftyReg, VoxelMorph and Ours (Feat.) on the Hippocampus dataset. The images in the first column are test image and moving images from VoxelMorph and NeurReg. The columns from second row to the last row are segmentation results from ANTs, NiftyReg, VoxelMorph and ours. The images in the last row are uncertainty map and MTL prediction backpropagated map to the moving image from VoxelMorph and Ours. The green and red region represents ground truth or true positive. Blue is the false positive region and yellow is the false negative region.}
\label{fig:seg_hippo}
\end{figure}

\begin{figure}[t]
\begin{center}
\begin{minipage}{0.19\linewidth}
	\includegraphics[width=\textwidth]{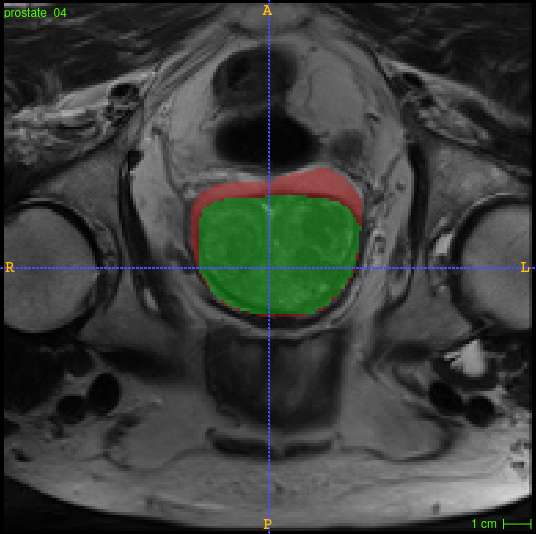}
\end{minipage}
\begin{minipage}{0.19\linewidth}
	\includegraphics[width=\textwidth]{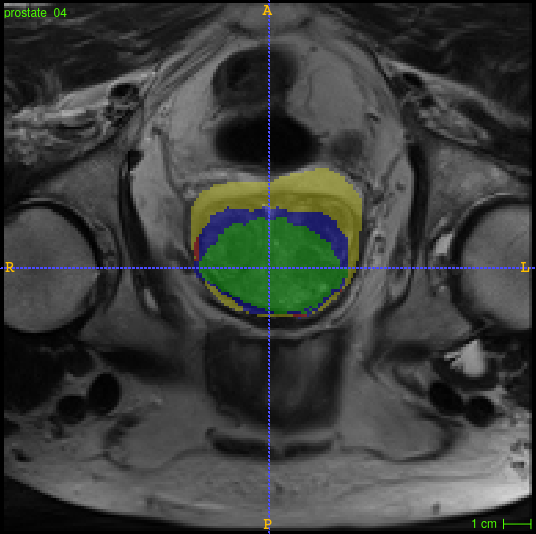}
\end{minipage}
\begin{minipage}{0.19\linewidth}
	\includegraphics[width=\textwidth]{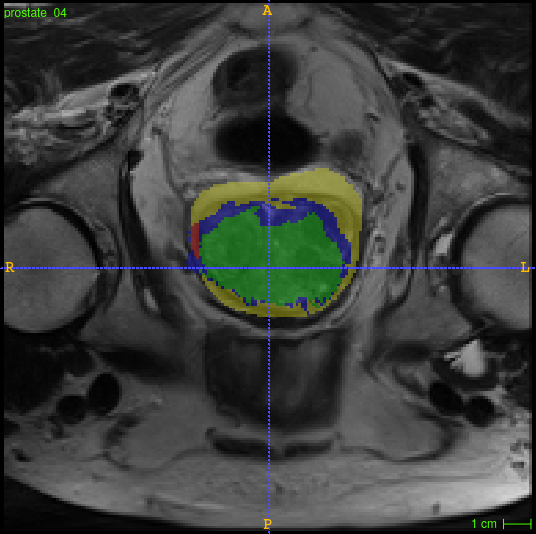}
\end{minipage}
\begin{minipage}{0.19\linewidth}
	\includegraphics[width=\textwidth]{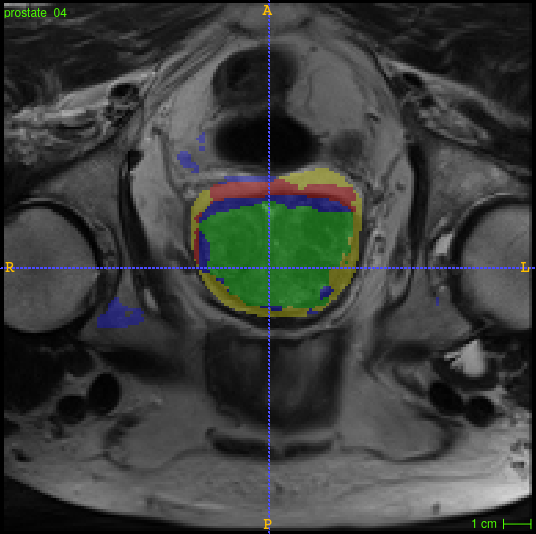}
\end{minipage}
\begin{minipage}{0.19\linewidth}
	\includegraphics[width=\textwidth]{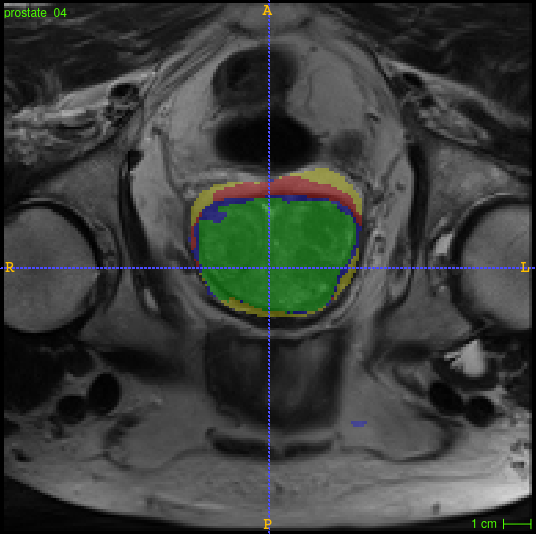}
\end{minipage}\\

\begin{minipage}{0.19\linewidth}
	\includegraphics[width=\textwidth]{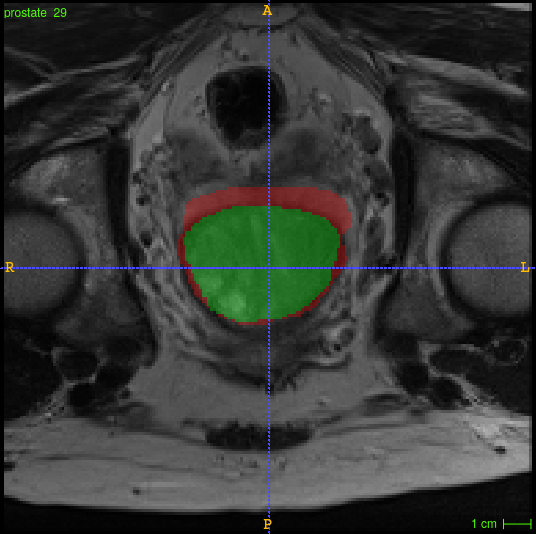}
\end{minipage}
\begin{minipage}{0.19\linewidth}
	\includegraphics[width=\textwidth]{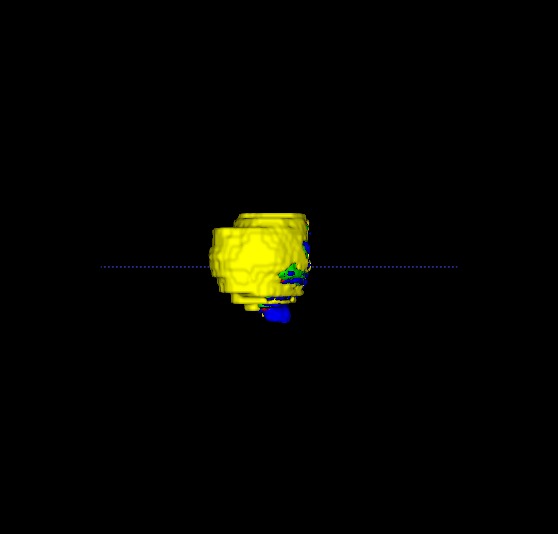}
\end{minipage}
\begin{minipage}{0.19\linewidth}
	\includegraphics[width=\textwidth]{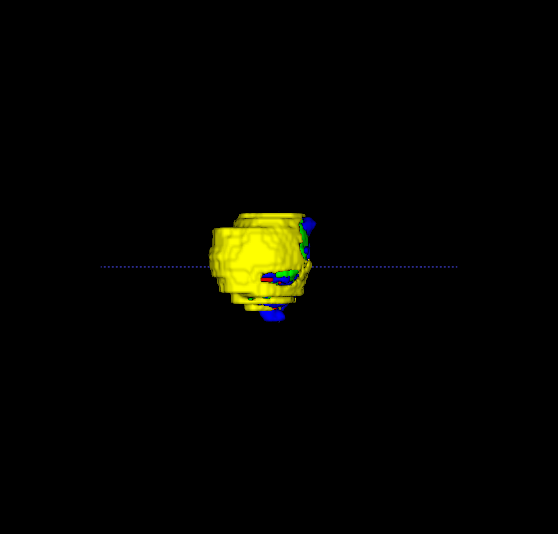}
\end{minipage}
\begin{minipage}{0.19\linewidth}
	\includegraphics[width=\textwidth]{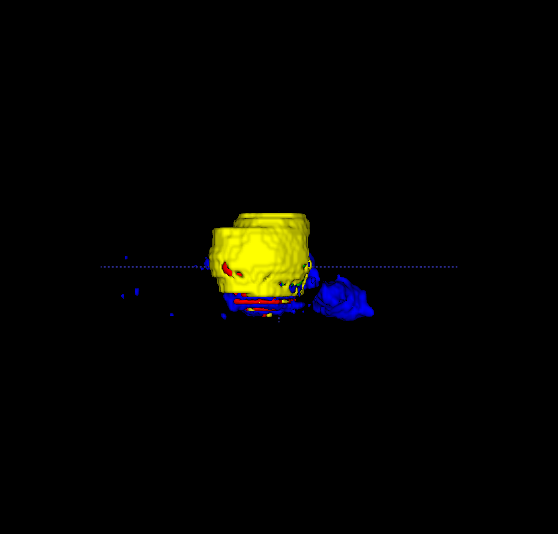}
\end{minipage}
\begin{minipage}{0.19\linewidth}
	\includegraphics[width=\textwidth]{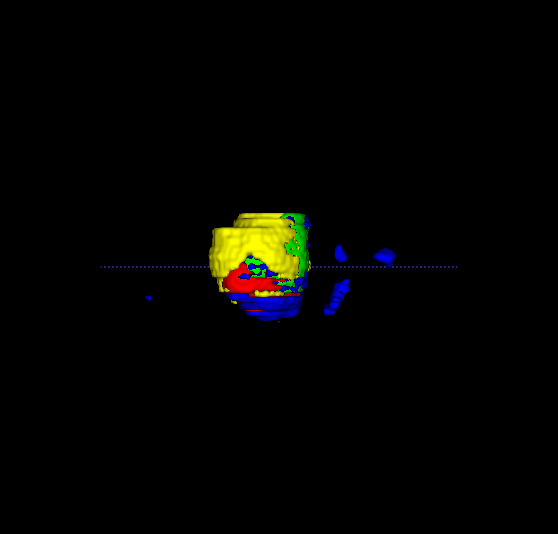}
\end{minipage}\\

\begin{minipage}{0.19\linewidth}
	\includegraphics[width=\textwidth]{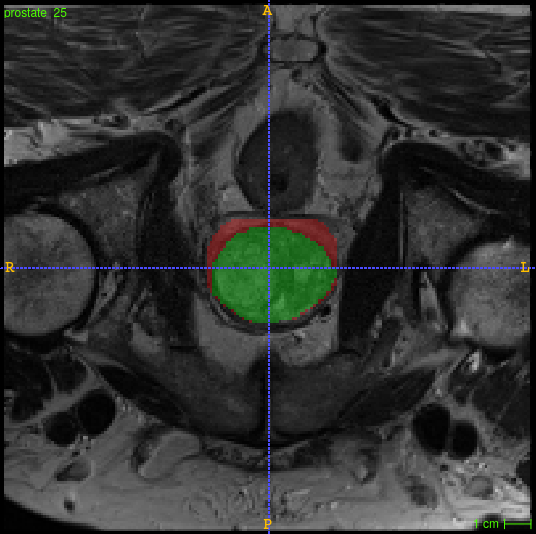}
\end{minipage}
\begin{minipage}{0.19\linewidth}
	\includegraphics[width=\textwidth]{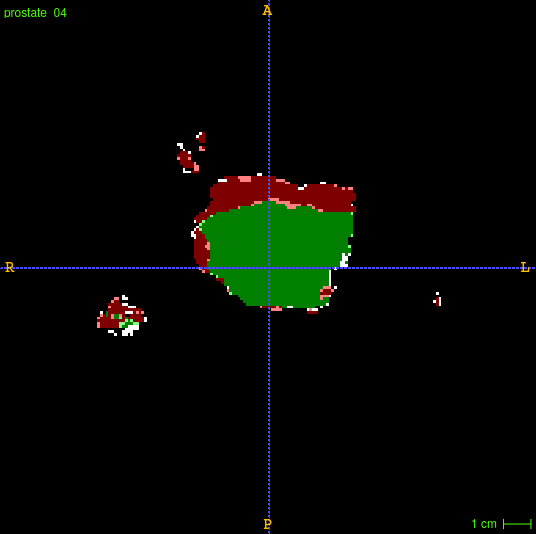}
\end{minipage}
\begin{minipage}{0.19\linewidth}
	\includegraphics[width=\textwidth]{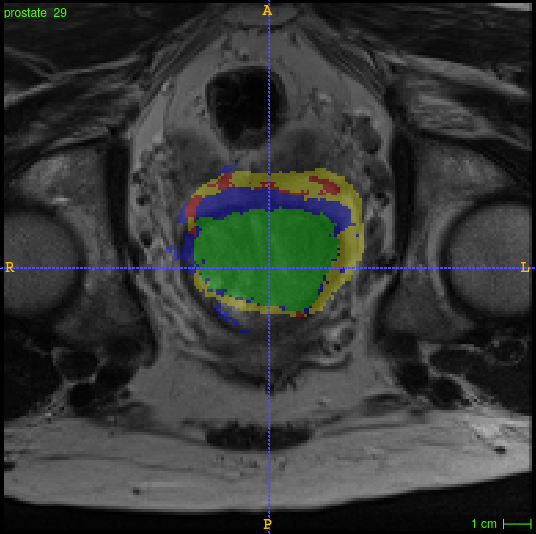}
\end{minipage}
\begin{minipage}{0.19\linewidth}
	\includegraphics[width=\textwidth]{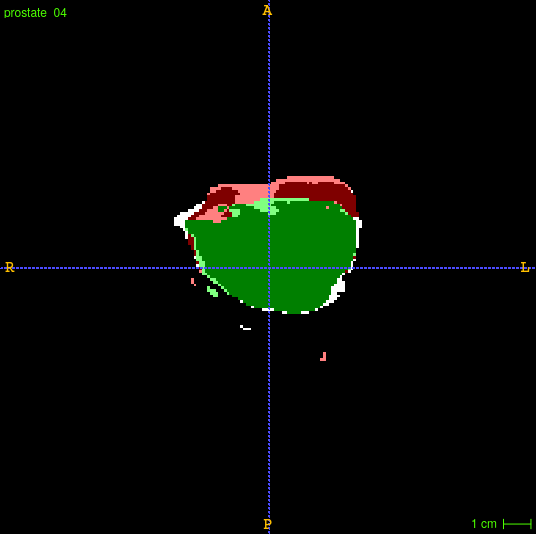}
\end{minipage}
\begin{minipage}{0.19\linewidth}
	\includegraphics[width=\textwidth]{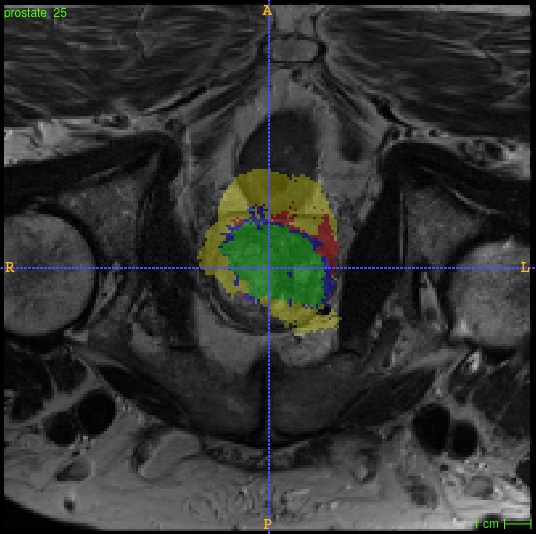}
\end{minipage}
\end{center}
   \caption{Visualization of segmentation results from ANTs, NiftyReg, VoxelMorph and Ours (Feat.) on the Prostate dataset.}
\label{fig:seg_pros}
\end{figure}

We visualize the segmentation from the four registration methods in Fig.~\ref{fig:seg_hippo} and~\ref{fig:seg_pros}. The figures demonstrate NeurReg achieves the best segmentation on the randomly chosen two test images from the two datasets.

\subsubsection{Uncertainty Estimation and Interpretability for Segmentation}
Uncertainty estimation is one of the most important features in medical image analysis to assist radiologists. For multi-atlas based segmentation, each atlas can be considered as a prior. We can simply derive the segmentation uncertainty by 
\begin{equation}\label{eq:uncert}
    \bm{U} = 1 - \frac{\sum_{i=1}^{N_{atlas}}{\mathbb{I}({\bm{S}_i=\bm{S}})}}{N_{atlas}},
\end{equation}
based on empirical estimation, where $\bm{S}$ is the prediction after majority voting and $\bm{S}_i$ is the prediction from $i_{th}$ atlas, $\mathbb{I}$ is an indicator image. If the pixel value in the uncertainty map is large, it means the prediction in the current pixel has a high uncertainty. From the uncertainty maps in last row of Fig.~\ref{fig:seg_hippo} and~\ref{fig:seg_pros}, the high uncertainty regions are along the edges of the predictions which are error prone areas. 

The other advantage of registration-based segmentation is the registration field provides the interpretability of segmentation prediction. From the registration field, we can build the connection between the MTL prediction and ground truth of the moving image. We use {ITK} to approximately calculate the inverse registration field in the last row of Fig.~\ref{fig:seg_hippo} and~\ref{fig:seg_pros}. Through the prediction backpropagated map based on top one image of NLCC, we can find the reason of the prediction even based on appearance in the image space. 
\section{Conclusion}
In this work, we developed a registration simulator to synthesize images under various plausible transformations. Then we design a hybrid loss between registration field supervision loss and data similarity loss in NeurReg. The registration field supervision provides an accurate field loss and is easy to optimize. The data similarity loss improves the model generalization ability. We further extend the registration framework to multi-task learning with segmentation and propose a dual registration to fully exploit the generalization of representational similarity loss on random fixed images. A residual segmentation block is designed to further boost the segmentation performance. Extensive experimental results demonstrate our NeurReg yields the best registration on several metrics and best segmentation with uncertainty and interpretability on the two public datasets. In future works, it is promising to generalize NeurReg and explore the applicability to multi-modal image registration.





{\small
\bibliographystyle{ieee}
\bibliography{egbib}
}

\end{document}